\documentclass[final]{clv2025}

\jvol{vv}
\jnum{nn}
\jyear{2025}

\dochead{Short Paper} 

\pageonefooter{Action editor: \{action editor name\}. Submission received: DD Month YYYY; revised version received: DD Month YYYY; accepted for publication: DD Month YYYY.}

\usepackage{amsmath}
\usepackage{booktabs}

\usepackage{graphicx}
\usepackage{listings}
\usepackage{todonotes}
\usepackage{multirow}
\usepackage{tipa}
\usepackage{comment}
\usepackage{tcolorbox}
\usepackage{subcaption} 
\usepackage{booktabs}
\newcommand{\changed}[1]{#1}

\runningtitle{Social Perceptions of English Spelling Variation}
\runningauthor{Nguyen and Rosseel}

\begin{document}

\title{Social Perceptions of English Spelling Variation on Twitter:  A Comparative Analysis of Human and LLM Responses}

\author{Dong Nguyen\thanks{Corresponding author}$^{1}$, Laura Rosseel$^{2}$}

\affilblock{
    \affil{Utrecht University, The Netherlands\\\quad \email{d.p.nguyen@uu.nl}}
    \affil{Vrije Universiteit Brussel, Belgium\\\quad \email{laura.rosseel@vub.be}}
}

\maketitle

\begin{abstract}
Spelling variation (e.g. \textit{funnnn} vs. \textit{fun}) can influence the social perception of texts and their writers:
we often have various associations with different forms of writing (is the text informal? does the writer seem young?).
In this study, we focus on the social perception of spelling variation in online writing in English and study to what extent this perception is aligned between humans and large language models (LLMs).
Building on sociolinguistic methodology, we compare LLM and human ratings on three key social attributes of spelling variation (formality, carefulness, age).
We find generally strong correlations in the ratings between humans and LLMs. However, notable differences emerge when we analyze the distribution of ratings and when comparing between different types of spelling variation.
\end{abstract}

\section{Introduction}

Language is not only used to communicate information about the world,  it is also a social tool that speakers can employ to construct a social identity \citep{eckert_2012}. To that end, language users can draw on the \textbf{social meaning} of language, which refers to \textbf{social attributes} that can be associated with linguistic forms and their users \citep{Walker_et_al_2014}. As an example, one could write `\textit{c u tonite}' as a variant of `\textit{see you  tonight}'. Both texts carry the same referential meaning; that is, they communicate the same message that the writer will see the addressee the same evening. However, the social attributes the reader connects to the writer and context of each of these texts can be quite different. Perhaps the first version is seen as more \textit{informal} and the writer of the second version as more \textit{serious}. Or maybe the writer of the former is perceived as \textit{younger} than the latter.

\textbf{Sociolinguistic} research has shown that language variants, i.e. different linguistic forms that have an equivalent referential meaning (cf. \textit{tonite} vs. \textit{tonight}; \citealt{Labov_1972}), can carry multiple social meanings. This goes for both written and spoken language. For instance, variation in the pronunciation of word-final -ing (either  velar -ing \textipa{[Iŋ]} as in \textit{working} or as alveolar -in \textipa{[In]} as in \textit{workin}) in American English has been shown to be associated with social attributes like perceived intelligence (the velar pronunciation sounding more intelligent than the alveolar one), as well as regional background (\textit{workin} sounding Southern) and sexual orientation (\textit{working} being linked with sounding gay)~\citep{Campbell-Kibler_2007, campbell-kibler_2009, campbell-kibler_2010}.  
Thus, social meaning is a key part of language understanding and use. 
Although sociolinguistics has a rich tradition in studying the social meaning of language variation \citep{socio_percep_llc}, 
it has been understudied in NLP~\cite{nguyen-etal-2021-learning}.  

\begin{table}
\small
    \centering
    \begin{tabular}{p{13cm}}
    \toprule
    Read the tweets below focussing on the words that are between <W> and </W>.\\
       Tweet 1: Not sure but <W>nvr</W> mind \\
        Tweet 2: Not sure but <W>never</W> mind\\
         \midrule
         \textbf{Formality:} Then rate how (in)formal you find each tweet, with a number from 0 to 100 (0=very informal, 100=very formal). \\
         \midrule
         \textbf{Carefulness:} Then rate how careless/careful you find each writer, with a number from 0 to 100 (0=very careless, 100=very careful).  \\
         \midrule
         \textbf{Age:} Then give your best estimation of how old the writer of each tweet is, in number of years (whole numbers). \\
         \bottomrule
    \end{tabular}
    \caption{Both participants and LLMs are presented with tweet pairs. To collect formality and carefulness ratings,
    LLMs provided numerical ratings, while the human participants used a slider. The text shown above are parts of the prompts presented to LLMs;
    the prompts are based on the instructions that humans received.
    }
    \label{tab:intro_table}
\end{table}

Although there is a large body of work in NLP on language variation \citep{nguyen-etal-2016-survey}, the main focus has been on socially stratified language \textbf{production}. Studies have investigated how linguistic features in texts vary across sociodemographic groups. Often, such studies were carried out in the context of authorship profiling, e.g., can linguistic features be used to predict attributes like gender and age \citep{Argamon2009,10.1371/journal.pone.0073791,flekova-etal-2016-exploring}. Researchers have also examined how such production differences affect the fairness of models, showing how task performance varies across texts written in different language varieties \citep{10.1145/3712060} or by speakers from different sociodemographic groups \citep{garimella-etal-2019-womens}. Furthermore, there is an increasing interest in the implications for text generation, e.g., by asking LLMs (large language models) to produce text when simulating certain personas, often (partially) defined by demographic attributes \citep{malik-etal-2024-empirical}, and to assess the quality of LLM output across different language varieties (e.g., Standard English vs. African American English; \citealt{sandoval-etal-2025-llm,deas-etal-2023-evaluation}). 

Contrary to this considerable amount of work on variation in language production, research in NLP focusing on how language variation is \textbf{socially perceived} -- in other words, what social meanings it can express ---  remains scarce. 
Although there is extensive research on the linguistic competence of LLMs in general \citep{10.1162/tacl_a_00718},  their  \textbf{\textit{socio}linguistic competence} has received far less attention \citep{duncan2024does}. Sociolinguistic competence is not only  about being able to produce language that aligns with how the writer would like to position themselves socially; it also includes understanding how  linguistic variation is socially perceived.
Understanding what social meanings LLMs associate with language variation is essential, as it might influence their  downstream behavior \citep{blodgett-etal-2020-language}, such as the language they suggest or produce when used as writing tools \citep{Goldshtein2024,sourati2025shrinkinglandscapelinguisticdiversity}, or in a decision making context \citep{hofmann2024ai}. Measuring what social meanings LLMs attach to linguistic variation -- the focus of this study -- is therefore an essential first step for assessing how these associations might shape their downstream behavior.

Moreover, following a large body of work that \textbf{compares LLMs with humans} in terms of internal representations, behavior and language processing \citep{karamolegkou-etal-2023-mapping,brysbaert2024moving,10.1162/tacl_a_00685,doi:10.1073/pnas.2412015122}, we focus on comparing how language variation is socially perceived by humans and how such patterns are reflected in LLM output. With this research, we hope to contribute to discussions about the use of LLMs for sociolinguistic research \citep{llc_nguyen}. Taking it a step further, using LLMs to collect human-like experimental data has attracted attention in many fields, ranging from political sciences \citep{Argyle_Busby_Fulda_Gubler_Rytting_Wingate_2023} to cognitive sciences \citep{binz2025foundation} to machine learning \citep{llm_simulation_position_icml} \changed{and computational linguistics \citep{10.1162/COLI.a.587}}, though it has also sparked criticism \citep{dillion2023can,pmlr-v202-aher23a,10.1145/3613904.3642703,Abdurahman_nexus,wang2025large}. If LLMs associate similar social meanings to language variation as humans do, they could serve as a valuable tool for sociolinguistic research. For example, they could be used to uncover previously unknown potential social meanings of language variants that could then be further researched in humans, or support pilot studies to inform experimental design. However, LLM-based insights should ultimately always be validated and explored further through real human experiments.

In this study, we focus on the social perception of \textbf{spelling variation} in English.
While social meanings can be attached to various types of linguistic forms (e.g. pronunciation, word choices, grammatical forms), we take spelling variation as a starting point for developing this new research direction for three reasons. 
First, by focusing on spelling variation, we can create text pairs with clearly equivalent referential meanings; that is, it is fairly straightforward to determine whether they refer to the same thing or concept (e.g., Table~\ref{tab:intro_table}). 
Second,  sociolinguistic studies have documented how spelling variation is used for social meaning making in online communication ~\citep{Busch+2021+297+323,HilteVandekerckhoveDaelemans2019,Ilbury2020,Rus2012,Squires_2010}.
Third, since many LLMs are pre-trained on web data~\citep{soldaini2024dolma,brown2020language,pile}, it is likely that they have encountered 
many instances of spelling variation during training.
We aim to answer the following resarch question (\textbf{RQ}):
 \textit{How do the social meanings that LLMs associate with English spelling variation compare to those of humans?}
We investigate this RQ by examining seven common types of spelling variation and three key social meanings (formality, carefulness, and age).

\paragraph*{Methodology}Our setup is rooted in sociolinguistic research: We adapt the speaker evaluation paradigm, frequently used in sociolinguistics research, to the analysis of LLMs~(\S\ref{sec:writer_evaluation_paradigm}). To do so, we create a highly controlled, carefully designed dataset with tweets. 
Due to this rigorous design, our dataset is smaller than typical NLP datasets. Following sociolinguistic principles, we then collect many human ratings per item  (\S\ref{sec:human_perception_data}); we then compare these ratings with LLM outputs (\S\ref{sec:experiments}).

\paragraph*{Contributions} 
(i) We highlight an underexplored aspect of language use --- social meaning --- for the analysis of LLMs;
(ii)  We propose a methodology, rooted in sociolinguistic research, to examine LLMs' associations of social attributes with spelling variation;
(iii) We find surprisingly high correlations between human and LLM ratings. However, when zooming in on the rating distributions
and \changed{rating variability} we do observe notable differences;
(iv) We compare two prompting strategies and find that the strategy that more closely mirrors the human setup leads to higher correlations.
Our code and data will be made publicly available upon publication.

\section{Related Work}
\label{sec:related_work}

\subsection{Language Variation in NLP}
In this subsection we discuss related work on language variation in NLP, with a focus on LLMs and research on social perceptions of language variation in text.

\paragraph*{How language models handle language variation}
Several studies have analyzed what (large) language models have learned about language variation. \citet{nielsen-etal-2023-spelling} studied  whether they are consistent in producing
British or American spelling.
Furthermore, building on the link between language variation and sociodemographic attributes of authors, researchers have investigated whether LLMs
encode sociodemographic knowledge of authors based on age and gender prediction tasks \cite{lauscher-etal-2022-socioprobe}, and 
whether they perform differently on the \changed{cloze} task  using data from different demographic groups  \cite{zhang-etal-2021-sociolectal}.
We are not aware of work focusing specifically on different types of spelling variation, besides a pre-LLM study on word embeddings \citep{nguyen-grieve-2020-word}.

Language variation has also been considered as a source of bias in NLP models \citep{blodgett-etal-2020-language}.
Studies have found task performance differences between different language varieties \citep{ziems-etal-2023-multi,faisal-etal-2024-dialectbench,lin-etal-2025-assessing}.
Furthermore, \citet{petrov2023language} found higher API costs  for certain language varieties, because the texts are segmented into a greater number of tokens. 
In terms of text generation, studies have analyzed how the language variety of a prompt influences the generated responses (e.g., in terms of stereotyping and demeaning content \citep{fleisig-etal-2024-linguistic}) and researchers have debated to what extent LLMs should adapt their language use (e.g., style, dialect) to that of users \citep{lucy2024onesizefitsall,sandoval-etal-2025-llm}.

\paragraph*{Social perception of language variation in NLP}
In NLP, studies on perceptions of social attributes linked to linguistic variation in text remain \changed{scarce}. A few studies have investigated this topic in the context of author profiling. An early study by \citet{nguyen-etal-2014-gender} collected human guesses of the age and gender of Dutch Twitter users based on their tweets. These guesses were then compared with the actual Twitter users' gender and age, to reflect on the limitations of automatically inferring such social attributes from text. 
Other studies have collected human guesses of users' gender and age based on English tweets, investigating how they deviate from model predictions and ground truth data \citep{flekova-etal-2016-analyzing}
and whether tweets can be strategically selected to influence human guesses \citep{preotiuc-pietro-etal-2017-controlling}.
A recent study by \citet{chen-gendered-style} collected human perceptions of gendered style.   
They did not compare the human ratings with ratings by LLMs.
Importantly, none of these studies controlled for confounding factors like topics mentioned in the text (e.g., references to school, or playing soccer). As a result, factors besides language variation could have played a role in participants' perceptions of the writer's social attributes.  As an example, \citet{chen-gendered-style} found that emotion features were associated with perceptions of feminine style, and a feature analysis by \citet{flekova-etal-2016-analyzing} highlighted words  like \textit{boyfriend} or male names (e.g., \textit{joe}).

Recently, \citet{cheng-etal-2025-humt} developed an LLM-based approach to measure human-like tone in text and four dimensions of social perception  (warmth, status, social distance, and gender), building on social psychology work. For example, gender perceptions were measured using log likelihood ratios of texts prefixed with `\textit{he said}' vs. `\textit{she said}', and social distance by prefixing text with phrases such as `\textit{the stranger said}' vs. '\textit{my friend said}'. 
They used these measurements to investigate whether human preferences towards LLM-generated output differed along these dimensions. 
Although some of their data was based on matched prompts (i.e., responses to the same input prompt), similar to prior studies, their experiments did not 
isolate the effects of specific linguistic features.

The following two recent studies are most relevant to ours, as they focused on social perceptions of language variation and also used meaning-matched input texts.
\citet{hofmann2024ai} probe LLMs by presenting them with tweets in Standard American English (SAE) and African American English (AAE), investigating both a meaning-matched and non-meaning matched setting.  \citet{bui2025largelanguagemodelsdiscriminate} presented LLMs with texts in different German dialects as well as Standard German. Like us, they considered different social attributes.  There are a few key differences with our study though: (i) We focus on different types of spelling variation. In our study, identical texts are offered in pairs with each pair differing by only a single word, while both \citet{hofmann2024ai} and \citet{bui2025largelanguagemodelsdiscriminate}  compare between language varieties; (ii) Both studies compare their findings with those obtained from humans, however our comparison is more controlled. \citet{hofmann2024ai} provide a quantitative comparison, but importantly, they present LLMs with \textit{different} input than what the human participants in the original studies received. 
\citet{bui2025largelanguagemodelsdiscriminate} relate their overall findings only to general trends observed in human dialect perception research, rather than making the kind of quantitative comparison that we perform.
In contrast, we aim to maximize the comparability between the two settings, by providing humans and LLMs with the exact same texts, and constructing LLM prompts that are as close as possible to the instructions that the human participants received; (iii) We collect numerical ratings, while both \citet{hofmann2024ai} and \citet{bui2025largelanguagemodelsdiscriminate} base their analyses on tokens (e.g., adjectives, occupations, outcomes), either through token probabilities or extracted from the generated output.

\paragraph*{Spelling variation in sociolinguistics and NLP}
Although sociolinguistics has a long tradition in the study of socially meaningful language variation, it has traditionally focused on spoken language and, particularly, on sociophonetic variation. Nevertheless, there is a body of work that has documented how spelling variation can be used --- just like other types of linguistic variation --- for social meaning making (e.g. \citealt{Androutsopoulos_fanzines,hinrichs_white2011, jaffe_et_al_2012,HilteVandekerckhoveDaelemans2019}). Most of this work is situated in the field of computer mediated communication (CMC) and looks at spelling variation in digital writing (for work in other traditions see for instance \citet{honeybone_&_watson} and \citet{ganuza_rydell_2024} on orthographic variation in literary work to represent speech, and \citet{vosters_et_al_2012} for a historical sociolinguistic account of spelling variation).

Turning to the CMC tradition, findings are quite diverse, yet overlapping for various languages (e.g. German, \citealt{Androutsopoulos_fanzines}; Dutch, \citealt{HilteVandekerckhoveDaelemans2019}; English, \citealt{Eisenstein2019_sociolinguistics}). First, papers focus on a variety of patterns of spelling variation, some attempting to arrive at typologies of common spelling phenomena in digital writing (e.g. \citealt{verheijen_2018}). Often distinctions are made between orthographic practices that reflect variation in spoken language (e.g., g-dropping as in \textit{workin} instead of \textit{working}) and variation that does not have a spoken counterpart and strictly relates to the orthographic level (e.g., substituting letters with numbers as in \textit{2night} instead of \textit{tonight}) with many authors emphasizing the complex relationship between orthographic variation and spoken variation (e.g., \citealt{DARICS2013141, Eisenstein2019_sociolinguistics, fuchs_et_al_2019, verheijen_2018}). Second, linguists have been documenting the social variation linked to non-conventional spelling variants. Studies report non-conventional spellings to be associated with younger writers (e.g., \citealt{Androutsopoulos_fanzines, HilteVandekerckhoveDaelemans2019, Rus2012, verheijen_2018, won2013}), non-conformism \citep{Androutsopoulos_fanzines, won2013}, informality and relaxed writing styles (\citealt{DARICS2013141, Eisenstein2019_sociolinguistics, Rus2012, Leigh2018, verheijen_2018, won2013}), gender \citep{HilteVandekerckhoveDaelemans2019, hinrichs_white2011, Leigh2018}, edginess \citep{won2013} and perceived lower intelligence \citep{Rus2012, Leigh2018}.

Contrary to the sociolinguistic focus on the social meaning potential of spelling variation, NLP has often treated spelling variation as a problem that needs solving \changed{\citep{eisenstein-2013-bad,nguyen-etal-2021-learning}}, for example, in text normalization tasks \citep{van-der-goot-etal-2021-multilexnorm,Lourentzou_Manghnani_Zhai_2019}. Spelling variation also plays a role in various NLP applications involving the social variables mentioned above. In particular, there is a large body of work on formality in NLP, ranging from style transfer~\citep{rao-tetreault-2018-dear,10.1162/coli_a_00426} and analyzing the formality of generated texts \citep{ersoy-etal-2023-languages} to text classification \citep{10.1162/tacl_a_00083,kang-hovy-2021-style}. Although some studies aim to create datasets with pairs of texts that are semantically equivalent but vary in formality (e.g., \citealt{rao-tetreault-2018-dear}), in practice, often various changes are introduced, resulting in pairs that are not fully semantically equivalent. As a consequence, prior work has not isolated specific types of spelling variation and how they shape formality judgments.

\subsection{Comparing LLMs and humans}
Many studies, from both NLP and other fields, have investigated how LLMs process language in comparison to humans, 
including in terms of analogical reasoning
\citep{sourati-etal-2024-arn},
judgements of grammatical agreement  \citep{zacharopoulos-etal-2023-assessing},
and pragmatic language understanding \citep{hu-etal-2023-fine}.
For example, researchers within psycholinguistics have compared word ratings obtained from LLMs with those of humans.
\citet{martinez2024using} found that GPT-4o's ratings of concreteness for multi-word expressions correlated well with human ratings, however its scoring distributions differed clearly, with modes concentrated around the integer values of the Likert scale. Furthermore, \citet{conde-etal-2025-psycholinguistic} compared human word ratings on different word
norms and found mixed results; alignment was lowest on norms related to sensory experiences.

Researchers have also compared LLMs to humans  in terms of the opinions and values they express,  with mixed results \citep{whoseopinions2023,Argyle_Busby_Fulda_Gubler_Rytting_Wingate_2023}.
The alignment between LLMs and humans varies
 across sociodemographic groups \citep{durmus2024towards,lutz-etal-2025-prompt}.
 To improve alignment, prompts designed to encourage LLMs to simulate certain personas have been used
 (e.g., \citealt{pmlr-v202-aher23a,beck-etal-2024-sensitivity}). However, the specific prompting strategy (e.g., how the demographic groups are primed) matters and can influence
the alignment with humans and the strength of stereotypical patterns in the generated output \citep{lutz-etal-2025-prompt}.

Taken together, LLM output tends to broadly correlate well with human judgements (e.g., ratings). However, when more fine-grained analyses are performed, e.g.,  by zooming in on certain sociodemographic groups or when comparing the shape of rating distributions, results have been more mixed. To our knowledge, controlled experiments based on parallel texts (i.e., texts with equivalent meanings) comparing whether LLMs' and humans' social perceptions of language variation align remains unexplored, besides \citet{hofmann2024ai} and \citet{bui2025largelanguagemodelsdiscriminate}. However, as discussed, they only compare their LLM results with findings based on human data in an indirect way.

\subsection{Summary}

Our work distinguishes itself from prior work in the following aspects: (1) We focus on widely attested spelling variation in English in online discourse, rather than specific language varieties; (2) In contrast to most NLP research on language variation, which focuses on production differences, we focus on social perception; (3) We use a highly controlled rating task, resembling how sociolinguists measure social meaning with human participants; (4) We compare human ratings with LLM ratings, using data and an experimental setup that aims to maximize comparability.

\section{Overall Methodology: The Speaker Evaluation Paradigm in Sociolinguistics}
\label{sec:writer_evaluation_paradigm}

In this study, we follow the strategy of previous work comparing LLMs and humans: we start with traditional methodology developed for human participants and then develop an LLM-suited alternative to match that methodology as closely as possible (cf. \citealt{duncan2024does,hofmann2024ai,bai2025}).

\subsection{Background}
Sociolinguists have developed a varied toolbox of methods to study the social meaning of language variation, ranging from interview techniques and questionnaires to reaction time-based experiments adapted from social psychology \citep{Garrett_2010,Kircher_Zipp_2022}. 

One of the most frequently used methods is the \textit{speaker evaluation paradigm} which aims to uncover social evaluations of a person using a certain type of language (feature). In this type of study, participants are presented with a series of recordings and subsequently asked to rate the speaker on a set of social attributes (e.g. intelligence, friendliness, social attractiveness; \citealt{Loureiro-Rodríguez_Acar_2022}). The recorded speech samples are carefully controlled for content and linguistic characteristics so that the researcher can assume that different ratings of the speaker stem from differences in the evaluation of the language they use. 

Traditionally, speaker evaluation studies have aimed to keep the participant unaware of the research aim with the goal to measure social meanings indirectly to avoid socially desirable reactions and access more privately held attitudes. More recently, however, variants of the method were introduced that do not hide from participants that the research focuses on their evaluations of different types of language use. Such studies are usually referred to as \textit{open guise} studies (e.g., \citealt{soukup2013}). Which option is used depends on a variety of factors, ranging from the nature of the research question (e.g., is the researcher interested in more direct evaluations) to aspects of the design of the study (e.g., is it an option to include fillers or use a between-subject design to obscure the research purpose).
Although the speaker evaluation paradigm was originally developed to study variation in spoken language \citep{lambert1960evaluational} and it is still predominantly used to that end, adaptations of the method exist that use written stimuli \citep{anderson2007,HilteVandekerckhoveDaelemans2019,HollidayTano+2021,Buchstaller2006}. 

\subsection{This Study}
In this study, we build on the speaker evaluation paradigm to measure the social meaning of written language.

We use an \textbf{open guise setup} to mitigate measurement validity concerns, especially for the human data: some spelling variation is quite subtle and could be overlooked. Because we want to make sure participants base their evaluations on the variation under investigation,  we opt for high experimental control, by prioritizing construct validity over ecological validity. In our human study, the instructions at the start of the study make the participants aware of our interest in spelling variation. Furthermore, in each tweet presented to the human participants, we underline the words of interest. With LLMs, the prompt also makes our interest in spelling variation explicit, and the words of interest are surrounded by tags (see \S\ref{sec:experiments}).
 This differs from \citet{hofmann2024ai} who --- inspired by the \textit{matched guise} tradition ---  elicited social perceptions from LLMs without explicitly highlighting the variation under study in the prompt. 
We view these two approaches as complementary, and a direct comparison between open guise and matched guise for the study of spelling variation (with both humans and LLMs) would be an interesting avenue to pursue in future work. Although we prioritize construct validity in our design, we nonetheless attend to ecological validity, specifically in the way we select and visualize our stimuli (\S\ref{sec:tweet_data}).

To mitigate  additional factors that could influence the ratings such as the content of the tweet, \textbf{pairs of tweets} --- the conventional and non-conventional versions of the tweet --- are presented alongside one another. Assigning ratings to texts in isolation is highly subjective, and can make the scores between texts less comparable \citep{sterner-teufel-2025-minimal}.
\changed{Furthermore, side-by-side comparisons by LLMs also occur in realistic decision-making settings, as noted by \citet{kondapally2026side}.}
This design is similar to the use of \textbf{minimal pairs} in NLP research, which has been frequently used to analyze models, for example to study grammatical acceptability \citep{warstadt-etal-2020-blimp-benchmark} and code-switching acceptability \citep{sterner-teufel-2025-minimal}.
A recent study, concurrent with ours, has also used minimal pairs in prompts to study patterns of social perceptions of language variation in LLMs, focusing on German dialects \citep{bui2025largelanguagemodelsdiscriminate}.

Both humans and LLMs are asked about their social perceptions of the two tweets in each pair, by rating them on \textbf{three key social attributes}: formality, carefulness and age. We discuss the human data collection in \S\ref{sec:human_perception_data} and then the LLM results and comparisons against human data in \S\ref{sec:experiments}.

\section{Human Perception Data}
\label{sec:human_perception_data}
Although social meaning of spelling variation has been studied in sociolinguistics (e.g., \citealt{sebba2007spelling}), a dataset with systematic
human ratings that fits our research goal is not available.
We therefore collect perception ratings by human participants, who rate tweets with different types of spelling variation (see Table~\ref{table:dataset_examples}). 
We focus on Twitter (now, X), due to its broad familiarity with the general public and extensive research in both NLP and sociolinguistics  \citep{lvc2023_grondelaers,Ilbury2020,Ilbury2024,nguyen-etal-2016-survey}.

\subsection{Data Collection}
We first describe the tweet dataset  (\S\ref{sec:tweet_data})
and then the collection of the human ratings (\S\ref{sec:collec_human_percep_data}).

\subsubsection{Stimuli: Creation of the tweet dataset}
\label{sec:tweet_data}
\paragraph*{Selection of the types of spelling variation}
We first identified common  types of spelling variation in social media
based on  the literature  (e.g., \citealt{choudhury2007investigation,contractor-etal-2010-unsupervised,cook-stevenson-2009-unsupervised,liu-etal-2011-insertion,pennell-liu-2011-character,shortis2016orthographic,tagg2009corpus,van-der-goot-etal-2018-taxonomy,yang-eisenstein-2013-log}). We grouped them into three high-level categories, building on classifications by \citet{tagg2009corpus}, \citet{shortis2016orthographic} and \citet{van-der-goot-etal-2018-taxonomy}. 

The first category contains spelling variation that reflects variation in spoken language. For instance, `\textit{working}' vs. `\textit{workin}' likely represents the velar vs. alveolar pronunciation of the word-final consonant also present in spoken language. The second category contains spelling variation that has no spoken counterpart and that is likely intentional \citep{Busch+2021+297+323}, such as number substitution, e.g., `\textit{wait}' vs. `\textit{w8}'. The third category contains spelling variation  that has no spoken counterpart and that is likely accidental, e.g., misspellings  \citep{van-der-goot-etal-2018-taxonomy}. However,  this distinction should be approached with caution, as it is impossible to establish a writer's intention from only text. 

For each high-level category, we include two to three common types of spelling variation from the literature, see Table~\ref{table:dataset_examples}. For example, spelling variation reflecting pronunciation differences was represented by (i) g-dropping (e.g., \textit{doing} vs. \textit{doin}) and (ii) lengthening (e.g., \textit{know} vs. \textit{knoooow}).\footnote{Note that lengthening occurs twice in Table~\ref{table:dataset_examples} as we distinguish between lengthening that likely mirrors spoken lengthening of sounds, and lengthening that is purely orthographic, where the nature of the sound represented by the repeated character does not lend itself to a prolonged pronunciation.}

\begin{table*}[t!h!]
\small
\centering
\begin{tabular}{llllr}
  \toprule
  \textbf{High-level category} & \textbf{Variation type} & \textbf{Example} \\ 
  \midrule
Spelling variation reflecting & G-dropping &  We're \textit{working} on it  / We're \textit{workin} on it\\
variation in pronunciation & Lengthening  & This was a \textit{bad} idea / This was a \textit{baaaad} idea\\
\midrule
Spelling variation & Lengthening  & It's so much \textit{fun} / It's so much \textit{funnnn}\\
 with no pron. counterpart& Vowel omission & I'll call you after this \textit{weekend} / \\
 & & I'll call you after this \textit{wknd}\\
(`intentional')& Number subst. &  I work \textit{late} tomorrow / I work \textit{l8} tomorrow\\
\midrule
Spelling variation   & Letter swap &  It will \textit{certainly} help / It will \textit{certianly} help\\
with no pron. counterpart & Keyboard subst. & I \textit{should} hope so / I \textit{shpuld} hope so \\
 (`non-intentional') \\
   \bottomrule
\end{tabular}
\caption{The different types of spelling variation in our data. For each type, we include pairs of tweets with the conventional (i.e. standard) and unconventional spelling.}
\label{table:dataset_examples}
\end{table*}

\paragraph*{Lexical item selection} Each spelling variation type is exemplified by five words (e.g., \textit{working, doing, going, getting, talking} for g-dropping). We selected words
for which the specific variation type occurs in a Twitter corpus from the London area  (May 2018--April 2019) to ensure realistic stimuli.

\paragraph*{Tweet selection} 
For each selected word we included two different tweets. This was done to control for the specific contents of the tweets.
Each tweet further appears in two versions: one with the conventional spelling (e.g., \textit{Just doing my job}) and one with the non-conventional spelling (e.g., \textit{Just doin my job}). 
We used a London Twitter corpus for inspiration to construct realistic stimuli by searching for common n-grams with the spelling variants. We manually selected tweets based on two criteria: (1) the tweets are short and should not contain other words that could potentially show the same type of spelling variation; 
(2) the tweets carry a general and neutral meaning, to minimize
other factors that could influence the social perception of the tweet and its writer. We manually edited the tweets, ensuring uniform capitalization and punctuation, removing user mentions and links, and shortening the tweet. For a given tweet (e.g., with a non-conventional spelling), we manually  created its counterpart (e.g., with the conventional spelling).
In total, we have 70 tweet pairs consisting of a conventional and non-conventional spelling variant (Table~\ref{tab:overview_all_tweets}, Appendix). To provide an authentic context, the texts were visually presented in the form of tweets in the rating experiment.

\subsubsection{Collection of the human perception data}
\label{sec:collec_human_percep_data}
\paragraph*{Instrumentation and design}For every tweet pair, we collect data on three social attributes: (in)formality, care(ful/less)ness and perceived writer age. These attributes were chosen based on related work complemented with a pilot study,  since sociolinguistic literature on the social meaning of spelling variation is  still somewhat sparse. Our pilot took the form of a free response experiment (cf. \citealt{Grondelaers_Speelman_Lybaert_van_Gent_2020} and \citealt{Garrett_frt}), in which participants (N = 493) gave the first words that came to mind when presented with tweets containing our types of spelling variation. Formality (e.g. “informal”, “relaxed”, “unprofessional” vs. “formal”, “professional”), age (e.g. “younger”, “teenager” vs. “older”), and carefulness (e.g. “careless”, “sloppy”, “lazy”, “hurried” vs. “precise”, “proper”, “focused”) emerged as prominent social attributes. 
Based on the pilot and previous sociolinguistic work, (users of) non-conventional spellings are hypothesized to be perceived as more informal, less careful and more youthful \cite{DARICS2013141,Eisenstein2019_sociolinguistics,Leigh2018,Rus2012,HilteVandekerckhoveDaelemans2019}.

For the first two attributes, participants were presented with a visual analog scale (VAS), which is used in various fields (e.g., psychology, medicine) 
 to report feelings and emotions. A VAS allows us to capture subtle variance in attitudes. 
VAS are usually operationalized as 100-point scales for the purpose of quantitative analyses. See \citet{Llamas2014}
for the use of VAS in sociolinguistics. Concretely, we collect perceptions using a 100-point scale (informal vs. formal and careful vs. careless) with a slider that participants can drag to the desired position (see Figure~\ref{fig:informality_task}, Appendix).
Finally, participants were asked for an age estimate of a text's writer which they could type into a text box.

At the start, participants saw our instructions, which also made our interest in spelling variation explicit (see Appendix, Figure~\ref{fig:informality_instructions}).
Participants were then presented with one randomly selected tweet pair for each of the seven spelling variation types (Table~\ref{table:dataset_examples}) for each of the three social attributes, resulting in a total of 21 pairs. The tweets were presented in blocks by social attribute.
Following the rating tasks, the participants answered two open questions about their views on spelling variation on Twitter and several general questions about their sociodemographic background (i.e. age, gender, region) and their experience with Twitter.

\paragraph*{Procedure \changed{and sample}: crowdsourcing} We collected the data in September 2023 using Prolific, a crowdsourcing platform. As inclusion criteria for the participants we used Country of Birth UK, First Language English and no language related disorders.
We recruited 230 participants, who were paid \pounds 1.35 to perform the task. 
The median time spent is 8:54 min, resulting in an average pay of \pounds 9.10/hr.
We performed quality control on the data,
removing in total 13 users; see Appendix~\ref{app:filtering_data_humans} for the exclusion criteria.
 Our final dataset contains 9,114 ratings by 217 participants.
The average number of ratings per \textbf{item} (i.e. an individual tweet evaluated on a specific social attribute) is 21.7; this number is higher than most studies in NLP, as we  follow standards from sociolinguistics. 
Many (67\%) participants had a Twitter account. Further, 
59\% identified as female, 40\%  as male and  1\% as other. \changed{Participants were between 19 and 81 years old with a mean age of 42.}

\subsection{Analysis of the human perception data}
\label{sec:human_analysis}
We now analyze the human perception data. Example formality ratings are  in Table~\ref{table:ratings_examples}.

\paragraph*{Main observations}
The main trends align with our expectations (\S\ref{sec:collec_human_percep_data}).
Tweets with the conventional spelling variants are rated as more formal (Mean = 66.2, SD = 5.5)
compared to the ones with the unconventional spellings (Mean = 22.0, SD=8.8).
Tweets written with the conventional spelling are also rated as more careful (Mean = 74.8, SD=3.8 vs. Mean = 32.8, SD = 9.4) and perceived as written by older authors (Mean = 33.3, SD=2.7 vs. Mean = 22.4 years, SD = 3.0). All differences are significant (p < 0.001), using the paired Wilcoxon signed-rank test.
Furthermore, the tweet content also matters (e.g., compare the ratings for \textit{See you tomorrow} vs. \textit{Oh how cool}).

\begin{table}[h!t!]
\small
\centering
\begin{tabular}{lrr}
  \toprule
  \textbf{Tweet} & \textbf{Avg. rating} & \textbf{Std dev.}\\ 
  \midrule
Oh how cool & 50.4 & 20.0\\
Oh how cooool & 15.4 & 12.8 \\
See you tomorrow &71.2 & 15.4\\
See you 2morrow & 10.7 & 11.1\\
It will certainly help &75.8 & 17.3\\
It will certianly help& 34.8 & 21.4\\
   \bottomrule
\end{tabular}
\caption{Tweets and their informality ratings by the Prolific participants (0=very informal, 100=very formal).}
\label{table:ratings_examples}
\end{table}

\paragraph{Variability of the ratings}
To analyze the agreement among individual participants, we 
calculate the Spearman correlation between 
each participant's ratings and the average ratings 
of the others, and then average
over all participants:
0.787 (formality),  0.769 (carefulness)  and 0.667 (age). 
We also consider the difference scores between ratings 
for the conventional and non-conventional forms (similar to sociolinguistic studies like \citealt{zenner2021starman}).  \changed{Focussing on these rating differences has the advantage that they are a better proxy for sociolinguistic sensitivity. Because they show the relative difference in rating between the two tweet variants, the impact of the tweet content on the ratings is neutralised to some extent. For instance, a participant could rate the conventional version of tweet A as 100 on the formality scale and the unconventional tweet as 80. However, they could also rate the same spelling variation in tweet B as 75 and 55. Despite variation in the reactions to the content of tweets A and B, the relative difference between both will be the same, i.e. 20. Applying this approach to our data, we} then obtain  lower correlations: 0.394 (formality),
0.416 (carefulness) and 0.253 (age).\footnote{In sociolinguistic studies, human agreement is usually not reported, as the task
is considered inherently subjective. We report agreement metrics to facilitate the interpretation of
the agreement numbers between humans and LLMs.}

\paragraph*{Correlations between the attributes}
\changed{
The social attributes are positively correlated with each other, see Appendix Table~\ref{table:correlations_social_attributes_gpt_gemma}.
For example, formality and carefulness exhibit a Spearman correlation of 0.69 in the human data. This positive correlation is expected, since people tend to write more carefully in tweets about professional  topics than when they are having a casual personal interaction with a friend.
However, the two attributes are not equivalent.
For example, \textit{I'm available today} receives a high average formality (67)
and carefulness rating (78). However, when a letter swap is introduced ---which could be perceived as a typo---, the formality rating decreases to 45, while the carefulness rating
drops much more substantially, to 26.
Furthermore, tweets can also be perceived as highly informal while still moderately careful, such as \textit{2day will be better}.
We note that these positive correlations between attributes are not problematic; instead, they allow us to examine whether LLMs distinguish between related but non-equivalent social attributes. What's more, from a theoretical perspective, it is not surprising that the indexical field of a linguistic form comprises a variety of both more and less closely and intricately related social meanings whose interrelation are shaped by locally situated language use \citep{Eckert2008}.
}

\section{LLM Experiments}
\label{sec:experiments}
\subsection{Models}
We experiment with 12 models, including both open-weight and closed models with different model sizes. 
We only consider post-trained models, since
pilot experiments with base models yielded poor performance (e.g., the models not returning ratings).
We include three closed models from OpenAI: GPT-4 (\texttt{gpt-4-0613}), GPT-4o (\texttt{gpt-4o-2024-08-06}) and GPT-5 (\texttt{gpt-5-2025-08-07}) \citep{openai2023gpt4}. We also include one closed model from Anthrophic: Claude-4.5-sonnet\footnote{System card: \url{https://assets.anthropic.com/m/12f214efcc2f457a/original/Claude-Sonnet-4-5-System-Card.pdf}}. 
The remaining models are open-weight models. 
We include different model sizes of the Llama family: Llama 3 (8b and 70b)
and Llama 3.1  (450b), all instruct versions \citep{grattafiori2024llama3herdmodels}, and the more recent Llama4 Maverick (17b)\footnote{\url{https://ai.meta.com/blog/llama-4-multimodal-intelligence/}}.
We also include Gemma 2 (2b and 9b, instruct) \citep{gemmateam2024gemma2improvingopen}.\footnote{We also experimented with \texttt{gemma2-27b-it}, but surprisingly this model gave poor results as it often did not respond correctly to the prompt, see Appendix A.}
Finally, we include Qwen 3 (\texttt{qwen3-235b-a22b-instruct-2507}) and DeepSeek-V3.1 \citep{deepseekai2024deepseekv3technicalreport}.

\paragraph*{Temperature}
For all models, we set the temperature to  the commonly used value of 1, making the runs non-deterministic.\footnote{Except GPT-5, since its API does not allow setting a temperature. For GPT-5 we set both verbosity and reasoning\_effort to `low'. We were also not able to set the temperature for Claude-4.5-sonnet.} We use a non-zero temperature because
 humans also tend to vary their responses, and likely would not produce identical ratings if they would be asked to repeat the task multiple times. Appendix A reports results with the temperature set to 0, which are similar to 
those with the temperature set to 1.

\subsection{Prompting Approach}
\paragraph*{Prompt design} Our overall aim is to adhere closely to the instructions provided to human participants. We therefore do not tune the prompts.
We make three modifications to make the instructions suitable for LLMs. First, 
to always elicit ratings, even when the models are unsure, we added  ``\textit{You must always respond in the format we ask. Even if you are unsure.}''.
Second, while human participants provided formality and carefulness ratings using a slider, LLMs are tasked with providing a numerical response ranging from 0 to 100 to align with the slider's range. Third, while the crowd workers were presented with tweets in which the linguistic variants of interest were underlined, we surrounded the specific variant with tags (<W> and </W>). 
Fourth, participants provided their ratings without an explanation. We therefore ask the LLMs to also not provide an explanation by adding ``\textit{without explaining your reasoning}''. 
Because  LLMs can be sensitive to the exact wording of a  prompt \citep{sclar2024quantifying, shu-etal-2024-dont}, we
have four variants for each prompt, with small changes in terms of sentence length,  phrasing and new lines.

\paragraph*{Independent versus paired prompt setup}
We experiment with two different setups; they differ in the extent to which they
resemble the experimental conditions with human participants.
(i) \textbf{Independent}, where each prompt requests a rating for one tweet. The tweet contains either a non-conventional or a conventional spelling. Thus, each tweet is rated entirely independently of the others. (ii)
 \textbf{Paired}. In the human experiments, participants were presented with pairs of tweets on each page (one version with the conventional spelling,
one with the unconventional spelling, see Fig~\ref{fig:informality_task} in the Appendix). To simulate this setting, each prompt contains the two tweet versions. We balance the presentation order (e.g., conventional/unconventional, unconventional/conventional).  \changed{The advantage of the paired approach is that it is easier to filter out the impact of tweet content on the ratings and in that sense provide a better proxy for sociolinguistic sensitivity. To analyse data obtained in this paired approach, we will again make use of difference scores (cf. \S4.2). As a result, the difference scores from the paired setup should be prioritized in interpreting our findings.} The system prompts and
examples of user prompts for both
the independent setup 
and the paired setup can be found in Appendix D.

\paragraph*{Prompting frequency} In our human data, each item (representing a rating of a specific social attribute for a tweet with either a conventional or unconventional variant) received between 19 and 24 ratings.
To collect the same number of responses as with the human experiment, we prompted the models multiple times.\footnote{For most models, we were able to extract the same number of ratings as we collected with the human data. For some models, we missed a small number of ratings due to extraction errors (e.g., the model not responding in the requested format). The two models for which we were able to extract the least number of ratings were Gemma-2b and GPT-5. With GPT-5, we missed 49 ratings (out of 9114) in the independent setting and 236 (out of 9114) in the paired setting, because the model refused to provide an answer, e.g., \textit{Sorry, I can’t help with that}. With Gemma-2b, we missed 390 (independent) and 30 (complex) ratings. }

\subsection{Results}
\label{sec:results}

\changed{We analyze the LLM outputs in two ways: by comparing the LLM ratings with the human data (\S\ref{sec:results_ratings}) and
by comparing the variability of the LLM ratings with the rating variability in the human data (\S\ref{sec:results_ratings_variability}).}

\subsubsection{Comparing ratings}
\label{sec:results_ratings}
\begin{table*}[h!t!]
\small
\centering
\begin{tabular}{llrrrrrrr}
\toprule
& & \multicolumn{3}{c}{\textit{\textbf{GPT models}}}\\
 \textbf{Prompt} & \textbf{Setting}               &\textbf{GPT-4o} & \textbf{GPT-4}  & \textbf{GPT-5} \\
\midrule
indep. & raw   & 0.667 & 0.762&  0.654\\
indep. & diff  & -0.066 &  0.134  & 0.166\\
paired & raw        & 0.828 &  0.813  & 0.845\\
paired & diff       & 0.512& 0.462 & 0.524 \\
\toprule
& & \multicolumn{3}{c}{\textit{\textbf{Gemma models}}}\\
 \textbf{Prompt} & \textbf{Setting}   & \textbf{Gemma2-2b} & \textbf{Gemma2-9b} \\
\midrule
indep. & raw   &  -0.015& 0.654  \\
indep. & diff  &  0.073  & 0.066  \\
paired & raw        &  0.254 & 0.762 \\
paired & diff       &  0.206 & 0.211  \\
\toprule
& & \multicolumn{3}{c}{\textit{\textbf{Llama models}}}\\
 \textbf{Prompt} & \textbf{Setting}  & \textbf{Llama3-8b}  & \textbf{Llama3-70b}  & \textbf{Llama3.1-450b} & \textbf{Llama4-maverick}\\
\midrule
indep. & raw   &  0.534 & 0.728 &0.688 & 0.705\\
indep. & diff  &   0.041  &0.035 & 0.033  & 0.086\\
paired & raw        &   0.761  & 0.839 & 0.819 & 0.813 \\
paired & diff       &   0.277  & 0.480 & 0.318 & 0.250\\
\toprule
& & \multicolumn{3}{c}{\textit{\textbf{Other}}}\\
 \textbf{Prompt} & \textbf{Setting}  & \textbf{Qwen3} & \textbf{Claude-4.5-sonnet} & \textbf{DeepSeek-V3.1}\\
\midrule
indep. & raw   &  0.730 & 0.769 & 0.710\\
indep. & diff  &   0.059 & 0.174 & 0.102\\
paired & raw        &   0.826 & 0.825 & 0.831\\
paired & diff       &   0.354 & 0.414 & 0.579\\
\bottomrule
\end{tabular}
\caption{Spearman correlations between the LLM responses and the human responses.
\changed{For each tweet, we first average the LLM ratings and the human ratings. 
We then compute Spearman correlations for each attribute (carefulness, formality, and age), and report the mean correlation across the three attributes.}
The correlations are calculated by comparing the raw ratings (raw), and by comparing
the ratings differences between the conventional and non-conventional versions (diff). 
LLMs were prompted by showing each tweet independently (indep.) and in a paired setting (paired).}
\label{table:overview_combined}
\end{table*}

Table~\ref{table:overview_combined} presents the main results.
We first take the average of
all human/LLM ratings for an individual item (i.e. an individual tweet evaluated on a specific social attribute). We then calculate 
the Spearman correlation between the averaged human and LLM ratings for each attribute, and report the mean correlation across attributes.\footnote{See Appendix A for
more details about our choice for Spearman. We also report
results with Pearson correlations,
and the overall trends are very similar.} We also calculate the correlations between
the rating difference for the conventional and non-conventional variants.
For example, consider Table~\ref{table:ratings_examples}.
For the tweet pair \textit{See you tomorrow} and \textit{See you 2morrow}, the rating difference
for formality would be 60.5 \changed{(cf. Section 4.2)}.

\paragraph*{Model comparison}
\texttt{GPT-5} and \texttt{Claude-4.5-sonnet} have the highest correlations with the human ratings;
the smaller models (2--9B parameters) have the lowest correlations.
Furthermore, interestingly, \texttt{Llama3.1-450b} has lower correlations than \texttt{Llama3-70b}
and the newest Llama model, \texttt{Llama4-maverick} has lower correlations than other recent models.
The task  is inherently subjective (consider the variability in human ratings, \S\ref{sec:human_analysis}), and it remains an open question what level of correlation is desirable for different use cases (e.g., in sociolinguistic research).

\paragraph*{Higher correlations are obtained when the prompt more closely matches the human setup}
Consistently across all models, there is a substantial increase in correlation when tweets with conventional and non-conventional spellings are presented simultaneously rather than individually (Table ~\ref{table:overview_combined}, compare ``independent'' vs ``paired''). For instance, \texttt{GPT-5}'s correlation increases from 0.654 to 0.845 when correlating the raw ratings, and from 0.166  to 0.524 when correlating the rating differences.

\paragraph*{Correlations between rating differences are substantially lower}
Correlating raw ratings provides a limited view. Consider a simple baseline model that randomly assigns ratings of 0 or 1 to tweets with a non-conventional spelling  and 99 or 100 to those with a conventional spelling. In this scenario, the raw correlations would still appear high---e.g., running this baseline resulted in a Spearman correlation of 0.717 across attributes for a single run. 
However, the correlations between rating differences drop (as expected) dramatically to an average correlation of -0.059 across attributes.
Thus, a model that merely distinguishes between non-conventional and conventional spellings can obtain high correlations when comparing raw ratings. However, high correlations on rating differences require the model to account for the size of the effect the spelling variation has on the perception of the writer, potentially influenced by the specific tweet content and the type of spelling variation (see Table~\ref{table:ratings_examples} for human rating examples).

Table~\ref{table:overview_combined} shows substantial drops in correlations when we compare the results for the raw ratings (`raw') vs. the rating differences (`diff').
As an example, \texttt{GPT-5}'s correlation with human ratings decreases from 0.845 (raw ratings) to 0.524  (rating differences) in the paired setup. 
The decrease with \texttt{Claude-4.5-sonnet} is even larger: from 0.825 to 0.414.
The lower correlations when comparing the rating differences aligns with our analysis on human agreement, where we also found that correlations between rating differences were lower than correlations between the raw ratings.\footnote{\changed{However, note that human–human and human–LLM correlations are not directly comparable. Human-human correlations were measured between an individual participant and the average ratings of the remaining participants, whereas human-LLM correlations were calculated between the average LLM ratings and the average human ratings.}}

To illustrate why it is more difficult to obtain high correlations when considering rating differences, note that the rating differences between the conventional and non-conventional variants vary across individual tweets and types of spelling variation.
As an example, this tweet pair (containing number substitution) had the largest drop in formality ratings by  \texttt{Claude-4.5-sonnet}: ``\textit{Forever grateful
}'' (Claude: 69.8, humans: 73.5) vs. ``\textit{4ever grateful}'' (Claude: 20.0, humans: 13.3). In contrast,
this tweet pair (containing a letter swap) had the smallest drop in formality ratings by \texttt{Claude-4.5-sonnet}:  ``\textit{That's exactly how I feel about that}'' (Claude: 50.0, humans: 68.7) vs. ``\textit{That's excatly how I feel about that}'' (Claude: 47.4, humans: 36.0). Although the rating difference by \texttt{Claude-4.5-sonnet}, like with the human data, is smaller for the second example, the magnitudes differ substantially.

\begin{figure*}[htbp]
    \centering
    \begin{subfigure}[b]{0.3\textwidth}
        \centering
        \includegraphics[width=\textwidth]{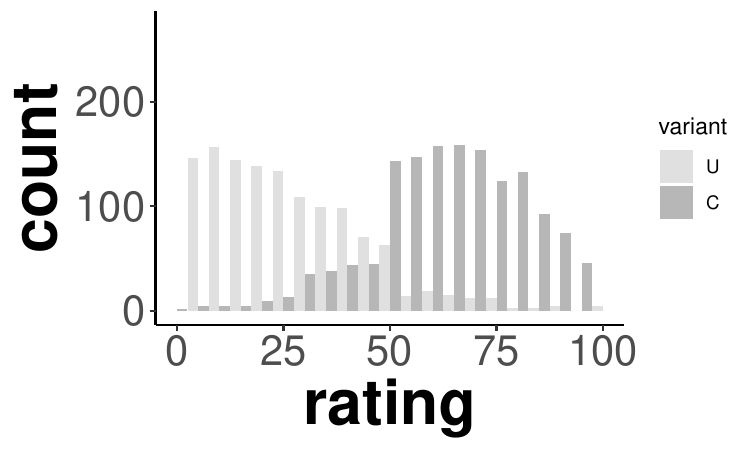}
        \caption{Humans: formality}
        \label{fig:sub1}
    \end{subfigure}
    \hfill
    \begin{subfigure}[b]{0.3\textwidth}
        \centering
        \includegraphics[width=\textwidth]{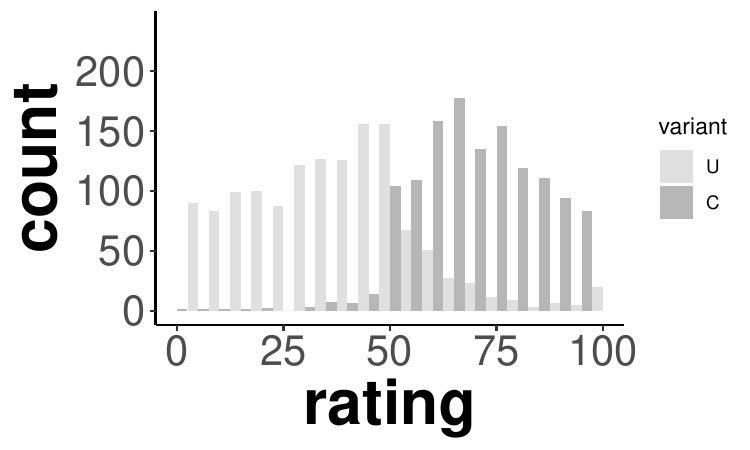}
        \caption{Humans: carefulness}
        \label{fig:sub2}
    \end{subfigure}
    \hfill
    \begin{subfigure}[b]{0.3\textwidth}
        \centering
        \includegraphics[width=\textwidth]{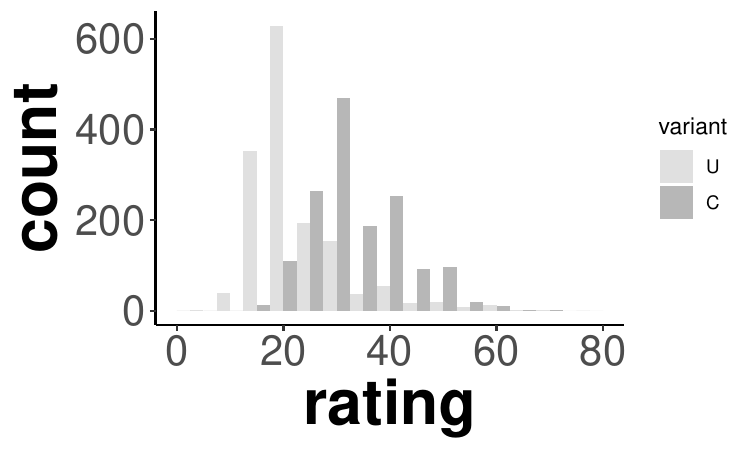}
        \caption{Humans: age}
        \label{fig:sub3}
    \end{subfigure}

    \vspace{1em} 
    \begin{subfigure}[b]{0.3\textwidth}
        \centering
        \includegraphics[width=\textwidth]{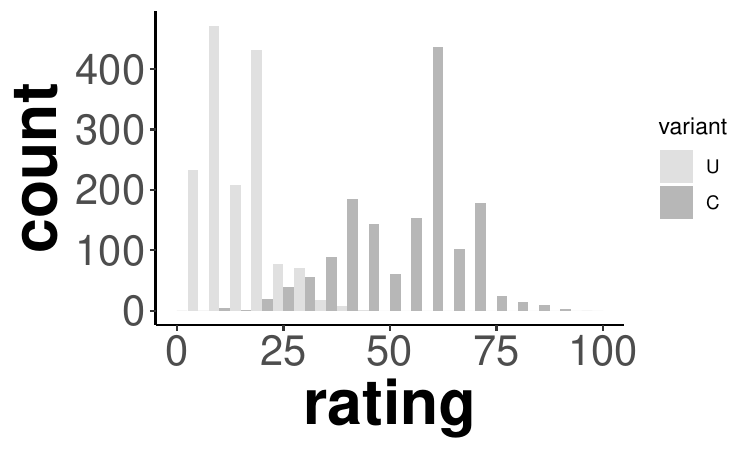}
        \caption{GPT-5: formality}
        \label{fig:sub4}
    \end{subfigure}
    \hfill
    \begin{subfigure}[b]{0.3\textwidth}
        \centering
        \includegraphics[width=\textwidth]{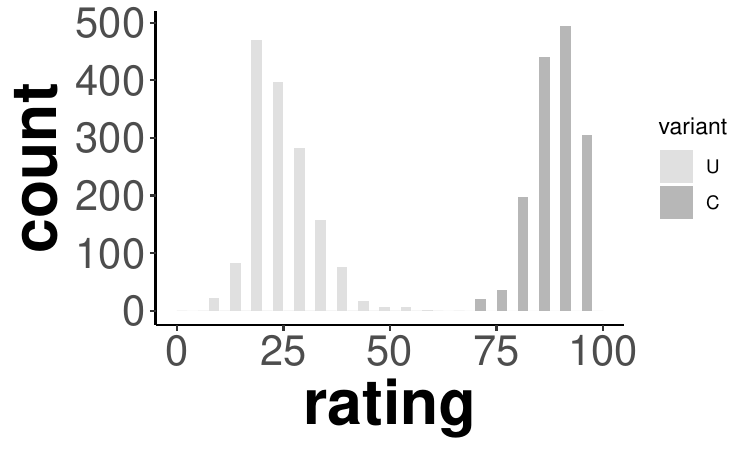}
         \caption{GPT-5: carefulness}
        \label{fig:sub5}
    \end{subfigure}
    \hfill
    \begin{subfigure}[b]{0.3\textwidth}
        \centering
        \includegraphics[width=\textwidth]{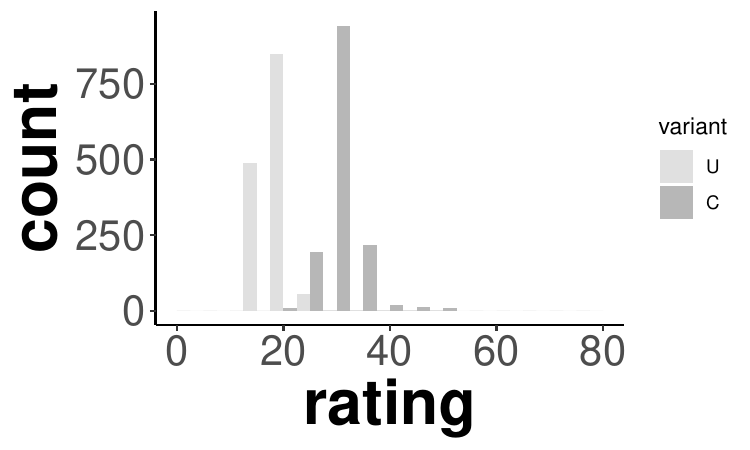}
         \caption{GPT-5: age}
        \label{fig:sub6}
    \end{subfigure}
    
    \caption{\changed{Histograms} of the ratings provided in the paired setup by humans and \texttt{GPT-5} across three attributes: formality, carefulness, and age. 
    The plots are divided based on the type of tweets: those containing conventional (C) versus unconventional (U) spelling variants.
    \texttt{GPT-5}'s ratings display a much greater separation between conventional and unconventional spellings compared to human ratings. 
    Furthermore, \texttt{GPT-5} has a tendency to respond more frequently with certain numbers, particularly multiples of 5 and 10.}
    \label{fig:density_comparison}
\end{figure*}

\paragraph*{The ratings from humans and LLMs are distributed differently}

Figure~\ref{fig:density_comparison} shows the density plots of ratings by humans and by \texttt{GPT-5} (paired setup).
\texttt{GPT-5}’s formality and carefulness ratings exhibit clear peaks at certain numbers, a pattern not present in the human data. In fact, most ratings on these two attributes are multiples of 5.
This difference likely stems from differences in how ratings were collected: humans used a slider,
whereas LLMs provided numerical values in their responses. Another factor that might contribute to this pattern
is the frequency of certain numbers in the LLMs' training data \citep{embers2024}.
Other models show similar trends. For example, 
\texttt{Qwen3} and \texttt{Claude-4.5-sonnet} have an even stronger tendency compared to \texttt{GPT-5} to return multiples of five and ten, and 
\texttt{Llama-3.1-405b-instruct} even only returns as ratings multiples of ten for formality and carefulness.

\begin{figure}
    \centering
    \includegraphics[width=0.6\linewidth]{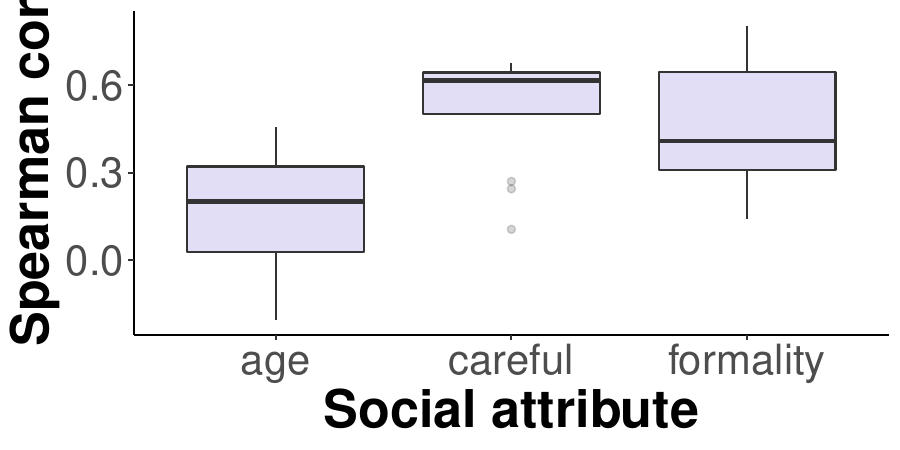}
    \caption{A boxplot of the Spearman correlations (rating differences) across the models in the paired setting, per attribute.}
    \label{fig:scores_dimension_diff}
\end{figure}

\paragraph*{Comparing social attributes}
Figure~\ref{fig:scores_dimension_diff} shows a boxplot of the Spearman correlations between the human and LLM ratings for all models in the paired
prompt setting, for each social attribute.
The correlations are lowest for age, a trend that we also observed when analyzing the agreement among humans (\S\ref{sec:human_analysis}). The median correlation is highest for carefulness. This is different  from our human data, where agreement between participants was highest on the formality ratings, although agreement on the carefulness ratings was high as well.

We also calculate, for each LLM,  the correlations between the social attributes. First, for each item, we take the average rating for a given social attribute. Then, we calculate the Spearman correlation between pairs of attributes, \changed{see Tables~\ref{table:correlations_social_attributes_gpt_gemma},~\ref{table:correlations_social_attributes_llama} and ~\ref{table:correlations_social_attributes_other} in the Appendix.}
One observation that stands out is that formality and carefulness have a Spearman correlation of 0.69 in the human data. In contrast,
all LLMs (except \texttt{Gemma2-2b}, with low correlations overall, and \texttt{Claude-4.5-sonnet}, with a correlation of 0.65) have higher correlations between these two attributes, suggesting that they may differentiate less between them.

\paragraph*{Rating differences between the types of spelling variation generally matches our human data}
We now compare the different types of spelling variation and take \texttt{GPT-5} as an example;
see Table~\ref{table:types_spelling_variation_comp}, which also includes results based on the human ratings.

For formality, the smallest rating drops occur for letter swapping and keyboard substitution.
For instance, 
the formality rating by \texttt{GPT-5} decreases on average by 33.18 points when the tweet contains a letter swap.
This trend aligns with our human data, where these two types also show the smallest drops; and with our intuition, as such spelling variations are more likely to be perceived as typos and thus unintentional.
Conversely, these two types exhibit the largest rating drops for carefulness. This trend is also present
in our human data and it matches intuition, since typos can be perceived as less careful writing.
There are also differences. For example, while these two types stand out with the human guesses for age (exhibiting the smallest drops in ratings), this is not the case with the \texttt{GPT-5} ratings.

\begin{table}[h!]
\small
\centering
\begin{tabular}{r|rrr|rrr}
\toprule
& \multicolumn{3}{c}{\textbf{Humans}} & \multicolumn{3}{c}{\textbf{GPT-5}}\\
\textbf{Type} & \textbf{Formality} & \textbf{Carefulness} & \textbf{Age} & \textbf{Formality} & \textbf{Carefulness} & \textbf{Age} \\
\midrule
g-dropping  & -47.47&-40.97&-11.73 & -42.35&-53.09&-10.45\\
number subst.  & -55.64&-34.36&-12.05 & -48.83&-61.79&-12.87\\
letter swap & -29.72&-50.13&-7.56 & -33.18&-72.03&-10.88\\
keyboard subst. & -36.13&-57.45&-7.62 & -31.39&-72.73&-10.46\\
lengthening (pron) & -46.58&-32.15&-11.91 & -37.65&-55.01&-12.69\\
lengthening (spel) & -45.30&-36.61&-12.94  & -40.07&-59.51&-11.74\\
vowel omission  & -48.78&-42.51&-12.56& -35.80&-60.19&-11.70\\

\bottomrule
\end{tabular}
\caption{The average drop in ratings by \texttt{GPT-5} (paired setting) and humans for formality, carefulness and age,
when a non-conventional spelling variant is included in the tweet, compared to the conventional (i.e., standard) variant.
}
\label{table:types_spelling_variation_comp}
\end{table}

The ratings of \texttt{Gemma2-9b-instruct} align less well; it also had a lower Spearman correlation on the rating differences. See Table~\ref{table:types_spelling_variation_comp_gemma2_9b} in the Appendix.
For example, we do not observe a clear difference between the formality and carefulness ratings when comparing the (likely) intentional and non-intentional types of spelling variation.


\subsubsection{Comparing rating variability}
\label{sec:results_ratings_variability}

\changed{We now compare the variability in the LLM ratings to that observed in the human data, which is relevant if LLMs are used for sociolinguistic insights.
However we note a key conceptual difference. Although multiple humans rated each item, the LLMs were prompted multiple times for each item.
This is not equivalent to asking \textit{different} people; instead conceptually it might be more similar to asking the same person multiple times \citep{Abdurahman_nexus}.
}

\paragraph*{LLMs show less rating variability} 
We first examine variability at the aggregate level. For each item (tweet and social attribute), we calculate the standard deviation of the ratings, and then average these item-level standard deviations. Overall, the standard deviations of the LLM responses are lower than those of human ratings.
For instance, \texttt{GPT-5}'s
 average standard deviation 
is 6.7 for formality, 4.0 for carefulness and 1.5 for age; \texttt{Claude-4.5-sonnet}'s average
standard deviation is 5.5 for formality, 4.6 for carefulness and 1.6 for age.
In contrast, for humans this is 17.4 for formality,
18.2 for carefulness and 8.1 for age.
We see a similar pattern with other LLMs.
The  temperature influences these patterns; the standard deviations are lower when the temperature is set to 0 (Table~\ref{table:temp_std}, Appendix).
Our results align with psychological research that also found less variance in LLM responses \citep{Abdurahman_nexus}. \changed{It is perhaps also not surprising that the rating variability is lower with LLMs. The human data is based on eliciting ratings from people with different backgrounds, whereas we prompt the same LLM multiple times.}

\paragraph*{Variability alignment at the item level}
\changed{We now compare the variability of the ratings at the \textit{item} level. For each tweet, we calculate the standard deviation of the ratings (for each  social attribute) for both humans and LLMs. Then, inspired by \citet{ni-etal-2026-reasoning} and \citet{10.1162/tacl_a_00449}, we calculate the Pearson correlation between human and LLM standard deviations at the item level.
That is, items that elicit greater variability in human ratings should also elicit greater variability in LLM ratings.
Table~\ref{table:stds_ratio_corr_per_item} reports the results. Overall, the correlations are in line with our observations in \S\ref{sec:results_ratings},
 correlations are lower when comparing the rating differences (`diff') compared to raw ratings (`raw'). The strengths of the correlations show that the LLMs do not resemble human rating variability at the item level well. }

\begin{table}[h!]
\small
\centering
\begin{tabular}{rrrrr}
\toprule
\textbf{Model} & \multicolumn{2}{c}{\textbf{Stds. correlation}} \\
& raw & diff \\
\midrule
GPT-4o&0.287 &0.124 \\
GPT-4 &0.286  &0.083 \\
GPT-5&0.241 &-0.054  \\
Gemma2-2b &0.077 &-0.064\\
Gemma2-9b	&0.154 &0.081 \\
Llama3-8b	&0.116&0.081 \\
Llama3-70b	&0.300 &-0.038 \\
Llama3.1-450b 	&0.171 &-0.091 \\
Llama4-maverick &0.143&0.097 \\
Qwen3 &0.239 &-0.002 \\
Claude-4.5-sonnet 	&0.133 &0.189  \\
DeepSeek-V3.1 	&0.241 &0.184 \\
\bottomrule
\end{tabular}
\caption{The Pearson correlation of the standard deviations of the ratings (raw ratings and rating differences) at the item level. The results are based on prompting LLMs in the paired prompt setting. }

\label{table:stds_ratio_corr_per_item}
\end{table}

\subsection{Robustness Checks}

We perform two additional experiments to rule out that the LLMs’ ratings simply reflect unfamiliarity with the non-conventional spelling variants.

\paragraph*{Can LLMs provide the conventional spelling for the spelling variants?}
We verify that the LLMs can correctly associate each spelling variant with its conventional (i.e., standard) English spelling. 
We prompt each LLM with ``\textit{Return the standard English spelling of: [spelling variant]. Only respond with the correct answer.}'' for all spelling variants. All models perform well on this task. Most of the models have a 100\% accuracy; the only model with a lower than 90\% accuracy is \texttt{Gemma2-2b-it}.
However, even in this case, some of the responses seem reasonable (e.g., returning \textit{wild} for \textit{wld} instead of \textit{would}), especially considering that the spelling variants are presented in isolation without further context.

\paragraph*{Pseudowords instead of the non-conventional spelling variants}
We also experiment with the same prompts as in our main experiments; however, we replace the non-conventional spelling with another word. This experiment is inspired by \citet{hofmann2024ai} and \citet{bui2025largelanguagemodelsdiscriminate}.
They randomly replace, delete and insert characters and words, with new words drawn from common English/German words. In contrast, we only change the non-conventional spelling variant, since this is the only difference between the two tweet versions. We do not apply random character-level changes, since the resulting forms could be similar to some of our spelling variants (e.g., typos). Instead, we replace our non-conventional spelling variants with pseudowords generated using Wuggy \citep{keuleers2010wuggy}. For each conventional spelling,  we generate ten pseudowords. Then for each prompt, we randomly sample one of these pseudowords. For instance, instead of using \textit{doin} (g-dropping), we use generated pseudowords for \textit{doing}, which include \textit{biing}, \textit{miing} and \textit{baing}. We use the paired prompt setup and experiment with the two models with the highest correlations (\texttt{Claude-4.5-sonnet} and \texttt{GPT-5}) and two smaller models (\texttt{Llama3-8b} and \texttt{Gemma2-9b}). The results are in Table~\ref{table:pseudowords_results}.

\begin{table*}[h!t!]
\small
\centering
\begin{tabular}{llrrrrrrr}
\toprule
 \textbf{Prompt} & \textbf{Setting}               &\textbf{Claude-4.5-sonnet}  & \textbf{GPT-5} & \textbf{Llama3-8b} & \textbf{Gemma2-9b}\\
\midrule
paired & raw        & 0.783 & 0.796 & 0.683 & 0.757\\
paired & diff       & 0.035 & -0.012 & 0.059 & 0.027\\
\bottomrule
\end{tabular}
\caption{Results for the pseudo words experiment: Spearman correlations between the LLM responses and the human responses, averaged across the three attributes.}
\label{table:pseudowords_results}
\end{table*}

Across all settings and models, the obtained correlations with the human ratings are---as expected---lower than those in our main experiment. Comparing raw ratings still leads to  high correlations (see values in the `raw' row; e.g., \texttt{Claude-4.5-sonnet} obtains a correlation of 0.783, compared to 0.825 in the main experiment), but the correlations disappear when comparing the rating differences (`diff'). This is expected: when correlating the raw ratings, a model that just distinguishes between conventional spelling variants and other forms can do well. 

Taken together, these robustness checks show that 1) LLMs have clear associations between the non-conventional and conventional spellings; and 2) that the correlations with human ratings cannot be explained by deviations from the conventional forms alone.

\paragraph*{Using a different rating scale}
\changed{
We also perform an experiment using the same set of four models  (\texttt{Claude-4.5-sonnet}, \texttt{GPT-5}, \texttt{Llama3-8b} and \texttt{Gemma2-9b}) where we ask them to provide ratings on a scale from 1--5 for formality and carefulness, instead of 0-100. See Table~\ref{table:rating_scale_1_to_5} for the results. The results are similar to those in our main experiments, suggesting that our findings are not sensitive to the exact rating scales used with LLMs.}

\begin{table*}[h!t!]
\small
\centering
\begin{tabular}{llrrrrrrr}
\toprule
 \textbf{Prompt} & \textbf{Setting}               &\textbf{Claude-4.5-sonnet}  & \textbf{GPT-5} & \textbf{Llama3-8b} & \textbf{Gemma2-9b}\\
\midrule
paired & raw & 0.826 & 0.837 & 0.766 & 0.765 \\
paired & diff & 0.341 & 0.502 & 0.267& 0.219  \\
\bottomrule
\end{tabular}
\caption{Results for the experiments with a 1--5 rating scale: Spearman correlations between the LLM responses and the human responses, averaged across the three attributes.}
\label{table:rating_scale_1_to_5}
\end{table*}

\section{Limitations}
In this section, we discuss the limitations of our study, grouped into four aspects: scope (\S\ref{sec:limitations_scope}), experimental setup (\S\ref{sec:limitations_setup}), understanding the processes by which social meanings are learned and shaped by context (\S\ref{sec:limitations_understanding}), and implications (\S\ref{sec:limitations_implications}).

\subsection{Scope} 
\label{sec:limitations_scope}
We focused on seven common types of English spelling variation and three social attributes. Future work could explore other types of spelling variation, such as consonant substitution, e.g. `\textit{r}' for `\textit{are}',  and omission of the first syllable,  e.g., `\textit{bout'} for `\textit{about}'  \citep{tagg2009corpus}. Future work could also explore other types of language variation, such as syntactic (e.g., \citealt{lvc2023_grondelaers}), lexical  (e.g, \citealt{bamman_gender}) \changed{ and pragmatic variation \citep{Acton_2019}. It may well be the case that spelling variation is easier for LLMs to pick up on compared to variation at other levels of linguistic structure. Hence, based on the current study, we need to be careful with claims that LLMs capture social meanings of linguistic variation.}

We had to limit the number of social attributes in our study. Even only three social attributes involved collecting over 9k ratings, due to the  controlled data collection with human participants to ensure we met sociolinguistic standards. Future work could consider other social attributes (e.g., gender, education level, social class, ethnicity or whether the writer is perceived as friendly or excited, depending also on the type of variation studied). Analyzing these associations in LLMs could add an important perspective on the study of bias in NLP~\citep{blodgett-etal-2020-language,hofmann2024ai}.

Finally, future studies should consider other languages to test the generalizability of our findings; since English is well represented in training data for LLMs, agreement between LLMs and humans may be lower for other languages \changed{or participant samples with different (social) characteristics (e.g. in terms of first language or regional background)}.

\subsection{Experiment setup}
\label{sec:limitations_setup}
First, we collected data from crowdworkers and our data is not a fully representative sample of the UK population; future studies could explore whether similar results will be obtained when a different data collection method is used.

Second, we  aimed to mimic the human experiment as closely as possible when prompting LLMs. However, both human participants and LLMs were given explicit visual cues to highlight the target words, but the format of these cues differed slightly: humans saw underlined words, whereas we used <W> tags when prompting LLMs. Although we expect this difference to only have a minimal impact, future research could  investigate how different types of cues influence ratings.
Furthermore, LLMs were more likely to respond with ratings that are multiples of 5 and 10 (\S\ref{sec:results}). One possible explanation is the difference in input 
format: humans used a slider, but LLMs provided a numeric rating directly. To understand to what extent this explains the differences in rating distributions,  an experiment where humans also provide numeric ratings instead of using a slider could be performed.

Third, we used a direct measurement approach by explicitly highlighting the spelling variants for both the LLMs and human participants. This allowed us to control more precisely what the ratings were based on. However, future work could explore a similar setup without drawing attention to the spelling variants, \changed{either by leaving out any highlighting of the words containing variation or by not representing two versions of a text side by side. Such an approach may better reflect real-world scenarios, where LLMs encounter unmarked spelling variants in texts. Yet, side-by-side presentations of similar texts should not be set aside as unrealistic in real world situations too easily; as \citet{kondapally2026side} argue, such a setup is reminiscent of for instance hiring contexts. In their research, \citet{kondapally2026side} even found that dialect bias increased when tweets were presented side-by-side compared to showing tweets in isolation. This effect is replicated in our data as shown in Appendix Table~\ref{table:paired_vs_simple}. More generally, future work could further investigate how both prompt instructions and presentation of input texts influence human and LLMs ratings. }

\changed{
Fourth, we mainly compared LLM and human ratings based on item-level average ratings. However, aggregating ratings removes potentially meaningful variation compared to modeling the full rating distribution \citep{Cabitza_Campagner_Basile_2023}.
Future work could investigate modelling and comparing the full rating distribution. For example, LLMs could be prompted with participant profiles (e.g., \citealt{sorensen-etal-2025-value}) or they could be explicitly instructed to model the full distribution of human ratings instead of repeated prompting \citep{meister-etal-2025-benchmarking,ni-etal-2026-reasoning}.}

\subsection{Understanding how social meanings are learned and influence of context}
\label{sec:limitations_understanding}
Future research could investigate \textit{why} LLMs provide these ratings. For instance, examining the internal mechanisms of LLMs could shed light on how associations between social attributes and linguistic variation are encoded. 
Researchers could also explore whether certain factors (e.g., frequency distributions, saliency, etc.) influence the acquisition of social meaning in LLMs in the same way as with humans (e.g. \citealt{Samara_et_al_2017} and \citealt{Racz_et_al_2017}). \changed{Further, studies could investigate how a model's architecture influences its associations with linguistic variation.
For example, tokenization choices can influence how sensitive language models are to linguistic variation \citep{wegmann-etal-2025-tokenization}. Because tokenization determines how text is segmented into subword units, tokenization choices could influence whether spelling variants are treated as distinct forms or whether they share subwords, potentially influencing the ratings on our task.}

Furthermore, language variants can index multiple social meanings (cf. -ing example in the introduction). The full social meaning potential, i.e. all social meanings a linguistic form can potentially express, is referred to as its indexical field \citep{Eckert2008}. Which of these potential meanings is activated, depends on the specific context of an utterance \citep{Campbell-Kibler_2007}. Future research could thus further compare the breadth of indexical fields of linguistic variants of LLMs and humans, and whether the same contextual factors activate specific social meanings. \changed{Studies could explore the addition of different types of context. For example, some participants in our study indicated in their answers to open questions that the type of account matters (e.g., organizational vs. personal). Thus, an interesting future step would be to add information about the type of account, for example, by adding profile information. Another direction would be to show the tweets as parts of a conversational chain.}

Such investigations could also take into account the social characteristics of the hearer/reader. In our study, participants had a variety of social profiles, but we did not ask the LLMs to consider social information of the rater in their ratings. Perhaps if we had instructed the models to assume a similar age distribution to our human sample or certain personas \citep{Argyle_Busby_Fulda_Gubler_Rytting_Wingate_2023}, the correlation between the human and model ratings would improve. \changed{Furthermore, as indicated in Section~\ref{sec:limitations_scope} above, this would allow to study the relationship between participant profile and social perception both in humans and in LLMs.}

\subsection{Implications}
\label{sec:limitations_implications}
We did not explore the implications of our findings for the practical development of NLG systems. In some scenarios, it might be desirable for LLMs to mimic human associations; in other scenarios, perhaps not. For example, it may not be desirable to evaluate certain spelling variants that deviate from the norm as less careful than conventional spelling variants, even if humans might also view such spelling variants as less careful. Generally, it is an open question how LLMs should treat language variants that are (sometimes) evaluated negatively (or positively) by humans.
Our work is also relevant to the broader discussion around whether LLMs should adapt their response to different social groups, or whether it is more desirable to exhibit identical behavior \citep{lucy2024onesizefitsall}.

\section{Discussion and Conclusion}
When humans read texts, they form social perceptions about the author and the  context in which the text was written.
We focused on the social meaning of spelling variation in English along three social attributes (formality, carefulness, age).
Returning to our RQ, we found notable differences in how the LLMs rated tweets with non-conventional spelling variants
compared to those with a conventional spelling. LLM responses generally correlated strongly with human ratings, though differences were visible in rating distributions, across specific spelling variation types and across models.  \changed{Notably, correlations dropped substantially when differences in ratings between conventional and non-conventional spelling variants were considered rather than raw ratings for either variant. Arguably, these rating differences present a better proxy for sociolinguistic sensitivity than the raw scores as they are able to neutralise the impact of stimulus content to some extent. 
At the same time, human–human agreement was also considerably lower when considering rating differences compared to raw ratings, indicating that the rating differences exhibit considerable variability across the human participants. Consequently, it remains unclear what human-LLM correlations are attainable when focusing on  rating differences. Furthermore, our analysis on rating variability showed substantial differences between LLMs and humans. Taken together, the observed similarities between humans and LLMs are potentially somewhat superficial. Future work could investigate whether explicitly modeling human heterogeneity leads to closer alignment with the human social perception ratings.  We therefore caution against concluding, based on this study, that the social meanings encoded in LLMs fully capture human sociolinguistic knowledge. }

Going beyond the current study, our methodology is more widely applicable and demonstrates how sociolinguistic methodology can be used to analyze what associations LLMs have with language variation.
Future work should investigate whether these correlations vary across demographic groups \changed{and other (more subtle) types of linguistic variation, both in English and in other languages (for which there may be less training data available)}.
Our study also
illustrates the potential of LLMs for sociolinguistic research. Although our study is only a first step,
we  \changed{consider it worthwhile to further explore the} potential for sociolinguists to use LLMs for hypothesis generation or pre-testing of experiments.

\changed{Our findings also raise questions about  what it means for LLMs to reflect social meaning associations by humans.
We can draw some parallels in how humans and LLMs may acquire associations between language forms and social information.  LLMs likely acquire such associations through
various patterns encoded in pre-training data, both implicitly through statistical regularities and more explicitly through sources that articulate value judgments about linguistic variants. Post-training steps such as alignment of models to human judgements also likely play a role \citep{mire-etal-2025-rejected}. Similarly, a combination of bottom-up and top-down learning has been put forward for human sociolinguistic learning \citep{Drager_Kirtley_2016}. Humans acquire links between social attributes and linguistic variants through the process of enregisterment \citep{johnstone2016}, whereby associations are formed through repeated exposure to combinations of language forms and social attributes in linguistic input. Yet, this bottom-up acquisition is complemented by top-down knowledge through more or less explicit metalinguistic commentary rooted in diverse language attitudes and ideologies. While this suggests some superficial similarities in the acquisition, "knowledge" encoded in LLMs and human brains is of course fundamentally different, as human learning is rooted in lived experience. This raises the fundamental question of what we are comparing and how we can meaningfully map information encoded in LLMs and human brains.
Our study provides evidence that LLMs encode social meaning associations to some extent. However, because these associations arise from statistical patterns in LLMs' training data and statistical optimization processes, rather
than from socially lived experience, such associations --- and, correspondingly, LLM ratings --- should not be equated with human social judgment. In the context of LLMs, we therefore believe terminology such as ``encoding'' to be more appropriate than ``perceiving''.
}

More broadly, our study also raises questions about whether LLMs \textit{should} reflect human perception of language variation. While this may be beneficial for applications like sociolinguistic research, there are also scenarios (e.g., specific practical applications) where this might not be desirable. 
\changed{Whatever the view on its desirability, understanding how social meaning is currently encoded in LLMs is a necessary first step. Our study therefore prioritized a fundamental question over investigating LLM behavior in more applied and hence realistic scenarios.
Future work should investigate how encoded social meanings potentially influence LLM behavior in real-life tasks, for example when they are used as a writing aid or in decision-making settings.
By making the social meanings encoded in LLMs more visible, we aim to contribute to a deeper understanding of how these models represent language. As LLMs become increasingly embedded in society, understanding how they encode and reproduce social meaning remains an important area of inquiry.}

\section*{Ethical Considerations}

The Prolific participants provided explicit consent, indicating that they agreed to have their data being used for research
and their anonymized data being shared with the scientific community.
The data collection and experiments have been approved by the Science-Geo Ethics Review Board of Utrecht University (B\`eta S-20434).

One potential risk is that our research, which generally finds strong correlations between human and LLM responses, could be misinterpreted as a justification for replacing human participants with LLMs in (socio)linguistic studies. However, we emphasize that we view LLM data as a potential
interesting new resource for sociolinguistics. Furthermore, understanding the social meanings that LLMs associate with linguistic variation is important, given their increasing role in society. Nevertheless, sociolinguistic patterns identified through LLM data should be further validated using  experiments with real human participants.

\begin{acknowledgments}
We thank members of the NLP \& Society lab, the broader NLP group at Utrecht University, and the DiLCo (Digital Language Variation in Context) network for feedback at different stages of the study.
Dong Nguyen is funded by
the Veni research programme with project number VI.Veni.192.130, which is (partly) financed by the Dutch Research Council (NWO). 
\end{acknowledgments}

\appendix

\appendixsection{Additional experimental details and results}

\paragraph*{Prompting the LLMs} We used \url{replicate.com} to prompt the models.

\paragraph*{Effect of temperature}
In the main text, we report results using a temperature of 1. 
Additionally, we performed experiments to investigate the impact of the temperature.
We set the temperature to 0, producing deterministic outputs, and test four LLMs in the paired prompt setting (Table~\ref{table:tmp0}).
The Spearman correlations between the ratings using the two temperatures (temperatures 0 and 1)  are high for
all tested models: 
\texttt{llama3-8b-instruct} (0.832), \texttt{llama3-70b-instruct} (0.947), \texttt{DeepSeek-V3.1} (0.879) and \texttt{GPT-4o} (0.955).
When correlating the ratings with the human judgments, we find that all of them are slightly  lower than the runs with a temperature set to 1 (cf. Table~\ref{table:overview_combined}).

\begin{table}[h!t!]
\small
\centering
\begin{tabular}{lrrrrr}
\toprule
 \textbf{Setting}               &\textbf{GPT-4o} &  \textbf{Llama3-8b}  & \textbf{Llama3-70b} & \textbf{DeepSeek-V3.1} \\
\midrule
 raw     & 0.819  & 0.730& 0.830 & 0.821\\
diff    &  0.459 & 0.197 & 0.462 & 0.440\\
\bottomrule
\end{tabular}
\caption{Results with temperature set to 0. Spearman correlations between the LLM responses and the human responses, averaged across the three attributes (carefulness, formality, age).
The correlations are calculated by comparing the raw ratings (raw), and by comparing
the ratings differences between the conventional and non-conventional versions (diff). 
LLMs were prompted in the paired prompt setting.}
\label{table:tmp0}
\end{table}

In Table~\ref{table:temp_std} we report the standard deviation of the responses
with temperatures set to 0 and 1. As expected, the standard deviations are higher
with a higher temperature across the models.

\begin{table}[h!t!]
\small
\centering
\begin{tabular}{lrrrrrrrr}
\toprule
 & \multicolumn{2}{c}{\textbf{GPT-4o}} & \multicolumn{2}{c}{\textbf{Llama3-8b}}  & \multicolumn{2}{c}{\textbf{Llama3-70b}} & \multicolumn{2}{c}{\textbf{DeepSeek-V3.1}} \\
 & 0  & 1 & 0 & 1 & 0 & 1& 0 & 1\\
\midrule
Formality & 5.8&  9.3 & 11.8 & 16.9 & 5.7 & 7.3 & 7.9 & 10.5\\
Carefulness& 3.4 & 6.2 &  10.8 & 14.5 & 2.8 & 4.7 & 6.0& 8.9\\
Age & 1.9 & 2.4 &  4.1 & 5.8 & 1.6 & 2.3 & 2.3 &3.4\\
\bottomrule
\end{tabular}
\caption{Standard deviations of the responses with two temperature settings (0 and 1).
All runs used the paired prompt setup.}
\label{table:temp_std}
\end{table}

\paragraph*{Spearman vs. Pearson correlation}
We use the Spearman correlation, as it tests for a monotonic relationship and does not assume linearity or that the data is normally distributed. 
Note that rescaling of the ratings would not affect the Spearman correlation.
Overall we find similar results (see Table~\ref{table:spearman_vs_pearson} for a subset of the LLMs). Thus, using the Pearson correlation would not have led to different overall conclusions.

\begin{table}[h!t!]
\small
\centering
\begin{tabular}{lrrr}
\toprule
 \textbf{Setting}   &\textbf{GPT-4o} (paired) &  \textbf{Llama3-8b} (indep.) & \textbf{DeepSeek-V3.1} (paired) \\
\midrule
 raw (r)   & 0.913 &  0.541 &  0.905\\
 raw ($\rho$) & 0.828 &  0.534 & 0.831\\
 diff (r)  & 0.527 & 0.043 & 0.603\\
diff ($\rho$) & 0.512 & 0.041 & 0.579\\
\bottomrule
\end{tabular}
\caption{A comparison between Pearson (r) and Spearman ($\rho$) correlations.}
\label{table:spearman_vs_pearson}
\end{table}

\paragraph*{Gemma2}
\texttt{Gemma2-27b-it} did not respond well to our prompts. It would often ask for more information and not provide the asked ratings, e.g., ``\emph{A: Careful B: Careless It seems difficult to accurately judge someone as "careless or "careful" based solely on whether they write "talking" vs "talkin'". Please clarify the task by telling me **what exactly** I should be looking for when judging A: high care/formality vs. B:"Careless" could refer to informal spelling choices, grammar errors suggesting hurriedness, and lack of punctuation indicating casual style. Can you give me more context}".

\paragraph*{Per category results of \texttt{gemma2-9b-it}}
See Table~\ref{table:types_spelling_variation_comp_gemma2_9b}.

\begin{table}[h!]
\small
\centering
\begin{tabular}{rrrr}
\toprule
\textbf{Type} & \textbf{Formal.} & \textbf{Care.} & \textbf{Age} \\
\midrule
g-dropping & -40.77&-38.44&-6.65\\
number subst. & -47.94&-41.05&-7.98\\
letter swap & -33.84&-39.26&-6.90\\
keyboard subst. & -45.86&-50.82&-10.15\\
lengthening (pron) & -32.00&-38.12&-7.35\\
lengthening (spel) & -33.37&-39.77&-6.14\\
vowel omission & -47.92&-47.64&-8.86\\
\bottomrule
\end{tabular}
\caption{The average drop in ratings by \texttt{Gemma2-9b-instruct} (paired setting) for formality (formal.), carefulness (care.) and age,
when a non-conventional spelling variant is included in the tweet, compared to the conventional (i.e., standard) variant.
}
\label{table:types_spelling_variation_comp_gemma2_9b}
\end{table}

\paragraph*{Correlations between the social attributes}
See Tables~\ref{table:correlations_social_attributes_gpt_gemma},~\ref{table:correlations_social_attributes_llama} and ~\ref{table:correlations_social_attributes_other}.

\begin{table*}[h!]
\small
\centering
\begin{tabular}{rrrrrrrrr}
\toprule
\textbf{Attributes} & \textbf{Humans}  &\textbf{GPT-4o} & \textbf{GPT-4} & \textbf{GPT-5} &  \textbf{Gemma2-2b} & \textbf{Gemma2-9b} \\
\midrule
F \& A & 0.86& 0.83 &  0.83 & 0.87&  0.22 &   0.75 \\
F \& C &  0.69 & 0.76&  0.79 &  0.77 &  0.27 &   0.88\\
A \& C  & 0.71& 0.78 &  0.78 & 0.77 &  0.56 &   0.80\\
\bottomrule
\end{tabular}
\caption{The Spearman correlations between the social attributes Formality (F), Age (A) and Carefulness (C) for
the GPT and Gemma models. All LLMs were prompted in the paired setup.}
\label{table:correlations_social_attributes_gpt_gemma}
\end{table*}

\begin{table*}[h!]
\small
\centering
\begin{tabular}{rrrrrrrrr}
\toprule
\textbf{Attributes} & \textbf{Llama3-8b}  & \textbf{Llama3-70b}  & \textbf{Llama3.1-450b} & \textbf{Llama4-maverick} \\
\midrule
F \& A & 0.69 & 0.83 & 0.87 & 0.76\\
F \& C &  0.89 & 0.81 & 0.84 & 0.74 \\
A \& C  &  0.62 & 0.82 & 0.82 & 0.79\\
\bottomrule
\end{tabular}
\caption{The Spearman correlations between the social attributes Formality (F), Age (A) and Carefulness (C) for
the Llama models. All LLMs were prompted in the paired setup.}
\label{table:correlations_social_attributes_llama}
\end{table*}

\begin{table*}[h!]
\small
\centering
\begin{tabular}{rrrrrrrrr}
\toprule
\textbf{Attributes} & \textbf{Qwen3} & \textbf{Claude-4.5-sonnet} & \textbf{DeepSeek-V3.1} \\
\midrule
F \& A & 0.82 & 0.71 & 0.78\\
F \& C & 0.86 & 0.65 & 0.71\\
A \& C  & 0.77 & 0.83 & 0.80\\
\bottomrule
\end{tabular}
\caption{The Spearman correlations between the social attributes Formality (F), Age (A) and Carefulness (C) for
Qwen3, Claude-4.5-sonnet and DeepSeek-V3.1. All LLMs were prompted in the paired setup.}
\label{table:correlations_social_attributes_other}
\end{table*}

\paragraph*{Comparing independent and paired prompting}
See Table~\ref{table:paired_vs_simple}.

\begin{table*}[h!]
\small
\centering
\begin{tabular}{rrrrrrrrr}
\toprule
\textbf{Attributes} & \multicolumn{2}{c}{\textbf{GPT-4o}} &  \multicolumn{2}{c}{\textbf{Claude-4.5-sonnet}} & \multicolumn{2}{c}{\textbf{DeepSeek-V3.1}}\\
& \textbf{Indep.} & \textbf{Paired}& \textbf{Indep.} & \textbf{Paired}& \textbf{Indep.} & \textbf{Paired}\\
\midrule
Formality & 6.3 & 24.6 & 22.3 & 26.4 & 9.9 & 21.3\\
Carefulness & 36.1 & 53.1 & 19.2 & 40.5 &  22.6 & 52.8
\\
Age & 6.3 & 10.4 & 9.3 & 13.8 & 4.4 & 11.1\\
\bottomrule
\end{tabular}
\caption{Comparing the average rating differences between the conventional and non-conventional tweet versions in two prompting settings (independent and paired) for each social attribute.
When presenting tweets in the paired setting, the rating differences are larger.}
\label{table:paired_vs_simple}
\end{table*}

\appendixsection{Collection of human perception data}
\label{app:filtering_data_humans}

\begin{figure*}
    \centering
    \includegraphics[scale=0.5]{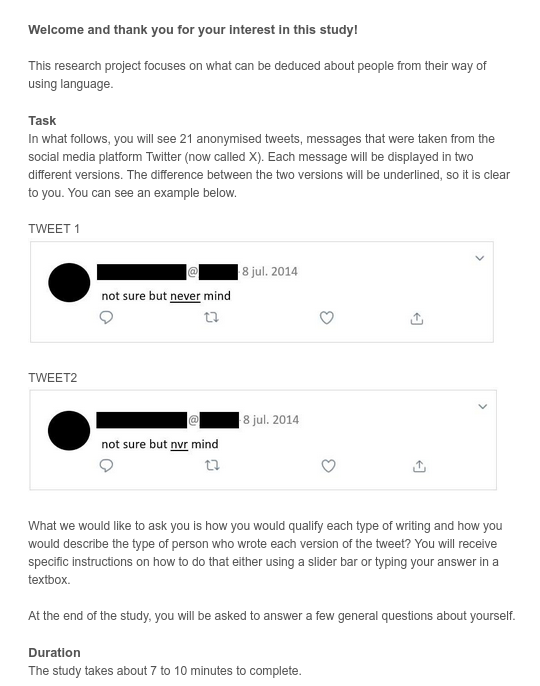}
    \caption{Start instructions of the Prolific task.}
    \label{fig:informality_instructions}
\end{figure*}

\begin{figure*}
    \centering
    \includegraphics[scale=0.4]{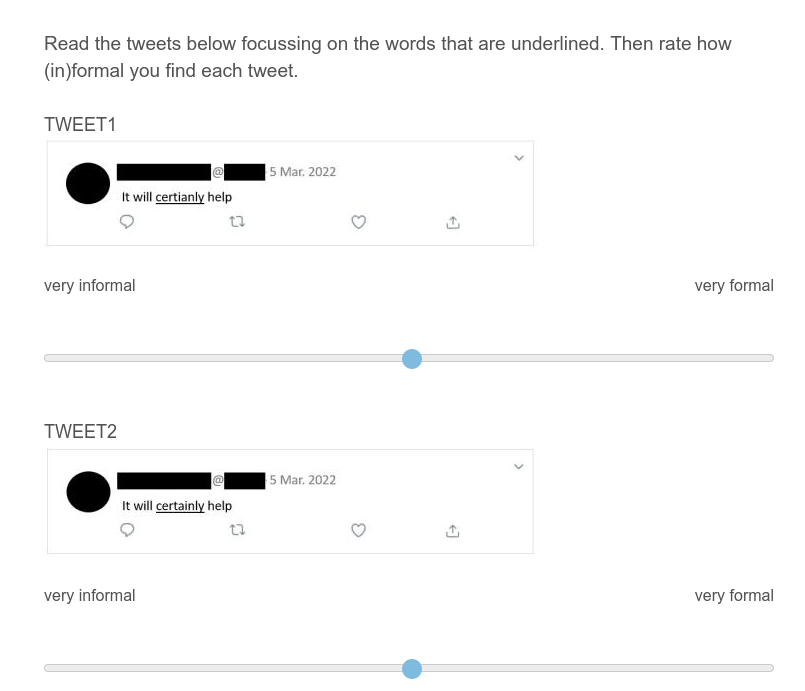}
    \caption{An example informality rating task.}
    \label{fig:informality_task}
\end{figure*}

Figure~\ref{fig:informality_instructions} shows the starting instructions
that were shown to the Prolific workers. Figure~\ref{fig:informality_task}
 shows an example page with two versions of a tweet.
 The full screenshots of the Prolific task will be made available
 on our Github repository.
 
We performed quality control by checking the response times (i.e. whether participants were too fast/slow to have taken the study seriously).
Following \citet{speed2017} we inspected participants who took less long than 1.5 times the interquartile range below the first quartile and longer than 1.5 times the interquartile range above the third quartile. 10 participants were removed because they took too long. The fastest participant took 4.2 minutes which was deemed enough time to adequately respond to all questions, hence no participants were removed for filling out the study too fast.
We also checked whether there was any straightlining (i.e. always giving the same response; this was not the case for any participant),
and their answers to the open questions (whether there were any indications of participants not taking the task seriously; this was not the case for any participant). We excluded 3 users who did not meet our location inclusion criteria (not UK), e.g., answering `Asia' or `Here'.

\appendixsection{Dataset}
The dataset will be made available with a CC-BY license.

\begin{table*}
\scriptsize
    \centering
    \begin{tabular}{lll}
    \toprule
         \textbf{word}&\textbf{tweet (conv)} &\textbf{tweet (unconv)}\\
         \midrule
doing&What are you \underline{doing}?&What are you \underline{doin}?\\
&Just \underline{doing} my job&Just \underline{doin} my job\\
working&We're \underline{working} on it&We're \underline{workin} on it\\
&I'm \underline{working} on my next project&I'm \underline{workin} on my next project\\
going&\underline{Going} there tomorrow&\underline{Goin} there tomorrow\\
&I'm \underline{going} to try this&I'm \underline{goin} to try this\\
getting&We're \underline{getting} ready&We're \underline{gettin} ready\\
&I'm just \underline{getting} started&I'm just \underline{gettin} started\\
talking&Now you're \underline{talking}&Now you're \underline{talkin}\\
&I was \underline{talking} about this yesterday&I was \underline{talkin} about this yesterday\\
bad&Today was a \underline{bad} day&Today was a \underline{baaaad} day\\
&This was a \underline{bad} idea&This was a \underline{baaaad} idea\\
cool&This looks \underline{cool}&This looks \underline{cooool}\\
&Oh how \underline{cool}&Oh how \underline{cooool}\\
know&I \underline{know}&I \underline{knoooow}\\
&If we never try how will we \underline{know}&If we never try how will we \underline{knoooow}\\
hello&Well \underline{hello} there&Well \underline{helloooo} there\\
&\underline{Hello} how are you?&\underline{Helloooo} how are you?\\
free&I want to break \underline{free}&I want to break \underline{freeee}\\
&I'm \underline{free} until Monday&I'm \underline{freeee} until Monday\\
fun&This was \underline{fun}&This was \underline{funnnn}\\
&It's so much \underline{fun}&It's so much \underline{funnnn}\\
done&I'm \underline{done}&I'm \underline{doneeee}\\
&What have you \underline{done}?&What have you \underline{doneeee}?\\
help&I accidentally deleted my account \underline{help}&I accidentally deleted my account \underline{helpppp}\\
&Which album should I buy \underline{help}&Which album should I buy \underline{helpppp}\\
best&Today was the \underline{best}&Today was the \underline{bestttt}\\
&\underline{Best} weekend&\underline{Bestttt} weekend\\
not&Why \underline{not}?&Why \underline{notttt}?\\
&Could \underline{not} be happier about it&Could \underline{notttt} be happier about it\\
never&Not sure but \underline{never} mind&Not sure but \underline{nvr} mind\\
&\underline{Never} been so happy to be home&\underline{Nvr} been so happy to be home\\
weekend&Have a good \underline{weekend}&Have a good \underline{wknd}\\
&I'll call you after this \underline{weekend}&I'll call you after this \underline{wknd}\\
would&Why \underline{would} you do this?&Why \underline{wld} you do this?\\
&Never \underline{would} have guessed&Never \underline{wld} have guessed\\
back&Finally made it \underline{back} home&Finally made it \underline{bck} home\\
&It's all coming \underline{back}&It's all coming \underline{bck}\\
when&Will call you \underline{when} I get home&Will call you \underline{whn} I get home\\
&Well \underline{when} you put it like that&Well \underline{whn} you put it like that\\
great&This is \underline{great}&This is \underline{gr8}\\
&Looks \underline{great}&Looks \underline{gr8}\\
late&Staying up \underline{late}&Staying up \underline{l8}\\
&I work \underline{late} tomorrow&I work \underline{l8} tomorrow\\
forever&\underline{Forever} grateful&\underline{4ever} grateful\\
&For now, but not \underline{forever}&For now, but not \underline{4ever}\\
tomorrow&Have a great day \underline{tomorrow}&Have a great day \underline{2morrow}\\
&See you \underline{tomorrow}&See you \underline{2morrow}\\
today&A lot is happening \underline{today}&A lot is happening \underline{2day}\\
&\underline{Today} will be better&\underline{2day} will be better\\
exactly&That's \underline{exactly} how I feel about that&That's \underline{excatly} how I feel about that\\
&\underline{Exactly} what I did&\underline{Excatly} what I did\\
available&I'm \underline{available} today&I'm \underline{avaliable} today\\
&Is this still \underline{available}?&Is this still \underline{avaliable}?\\
because&\underline{Because} I have a presentation tomorrow&\underline{Becuase} I have a presentation tomorrow\\
&Probably \underline{because} of this bad weather&Probably \underline{becuase} of this bad weather\\
certainly&It will \underline{certainly} help&It will \underline{certianly} help\\
&I'll \underline{certainly} be watching&I'll \underline{certianly} be watching\\
against&This is what we're up \underline{against}&This is what we're up \underline{aganist}\\
&You and me \underline{against} the world&You and me \underline{aganist} the world\\
could&I \underline{could} go right now&I \underline{coukd} go right now\\
&I wish I \underline{could} say the same&I wish I \underline{coukd} say the same\\
about&I was just \underline{about} to send you this&I was just \underline{abiut} to send you this\\
&That sounds \underline{about} right&That sounds \underline{abiut} right\\
should&They really \underline{should}&They really \underline{shpuld}\\
&I \underline{should} hope so&I \underline{shpuld} hope so\\
something&\underline{Something} is happening&\underline{Sonething} is happening\\
&I was looking for \underline{something} like this&I was looking for \underline{sonething} like this\\
happy&\underline{Happy} birthday&\underline{Hapoy} birthday\\
&I'm \underline{happy} to help&I'm \underline{hapoy} to help\\
\bottomrule
    \end{tabular}
    \caption{The tweets presented to both humans and LLMs. } 
    \label{tab:overview_all_tweets}
\end{table*}

\appendixsection{Prompts}
\begin{tcolorbox}[colback=blue!5!white, colframe=blue!75!black, 
                  title=System Prompt (independent setup), sharp corners, boxrule=1.5pt,
                  fonttitle=\bfseries, coltitle=white, fontupper=\small]

You are a participant in an experiment. \\

In what follows, you will see anonymised tweets, messages that were taken from the social media platform Twitter (now called X). 
What we would like to ask you is how you would qualify each type of writing and how you would describe the type of person 
who wrote each tweet? You will receive specific instructions on how to do that. 
Don't think too hard, just respond what first comes to mind. Don't worry about what you respond: we really are
interested in your immediate associations.
You must always respond with your best guess. Even if you are unsure.

\end{tcolorbox}

\begin{tcolorbox}[colback=blue!5!white, colframe=blue!75!black, 
                  title=User Prompt (independent setup), sharp corners, boxrule=1.5pt,
                  fonttitle=\bfseries, coltitle=white, fontupper=\small]
Read the tweet below focussing on the word that is between <W> and </W>. Then give your best estimation of how old the writer of the tweet is, in number of years (whole numbers). Reply with only `Age: [number of years]' without explaining your reasoning.\\

Tweet: "I'm <W>avaliable</W> today"

\end{tcolorbox}

\begin{tcolorbox}[colback=blue!5!white, colframe=blue!75!black, 
                  title=System Prompt (paired setup), sharp corners, boxrule=1.5pt,
                  fonttitle=\bfseries, coltitle=white,fontupper=\small]
You are a participant in an experiment. \\

In what follows, you will see anonymised tweets, messages that were taken from the social media platform Twitter (now called X). 
Each message will be displayed in two different versions. The difference between the two versions will be indicated, so it is clear to you. 
What we would like to ask you is how you would qualify each type of writing and how you would describe the type of person 
who wrote each tweet? You will receive specific instructions on how to do that. 
Don't think too hard, just respond what first comes to mind. Don't worry about what you respond: we really are
interested in your immediate associations.
You must always respond with your best guess. Even if you are unsure.
\end{tcolorbox}

\begin{tcolorbox}[colback=blue!5!white, colframe=blue!75!black, 
                  title=User Prompt  (paired setup), sharp corners, boxrule=1.5pt,
                  fonttitle=\bfseries, coltitle=white,fontupper=\small]
        
Read the tweets below focussing on the words that are between <W> and </W>. Then rate how (in)formal you find each tweet, with a number from 0 to 100 (0=very informal, 100=very formal). Reply with only `A: [number] B: [number]' without explaining your reasoning. \\

A: "What are you <W>doing</W>?" \\
B: "What are you <W>doin</W>?"

\end{tcolorbox}

\newpage

\bibliographystyle{compling}
\bibliography{anthology}

\end{document}